\documentclass{article}

\usepackage[preprint]{rhymes_2024}
\usepackage{palatino}
\usepackage{graphicx}
\usepackage{colortbl}
\usepackage{pgf}
\usepackage{subcaption}

\usepackage{xcolor} 

\definecolor{lightorange}{rgb}{1.0, 0.9, 1.0}  
\definecolor{lightgreen}{rgb}{0., 0.66, 0.}  

\newcommand{\rgb}[1]{%
  \pgfmathsetmacro{\percent}{%
    (max(min(#1, 80), 30) - 30) / (80 - 30) * 100
  }%
  \edef\temp{\noexpand\cellcolor{lightgreen!\percent!lightorange}}\temp}

\usepackage{amsmath}
\usepackage{bbm}
\usepackage{arydshln}
\usepackage{multirow}
\usepackage{enumitem}
\usepackage{wrapfig}
\usepackage{makecell}
\usepackage{pgfmath}
\usepackage{pgffor}
\usepackage{float} 
\usepackage{xspace}
\setlist[itemize]{leftmargin=*}

\definecolor{deepgreen}{RGB}{0,150,0} 
\definecolor{deepred}{RGB}{150,0,0} 

\usepackage{pifont}

\usepackage[utf8]{inputenc} 
\usepackage[T1]{fontenc}    
\usepackage[colorlinks=true, linkcolor=black, urlcolor=blue, citecolor=blue]{hyperref}
\usepackage{url}            
\usepackage{booktabs}       
\usepackage{amsfonts}       
\usepackage{nicefrac}       
\usepackage{microtype}      
\usepackage{xcolor}         
\usepackage[misc,geometry]{ifsym}

\usepackage{graphicx}
\usepackage{array}
\usepackage{amsmath}
\usepackage{listings}

\renewcommand{\ttfamily}{\fontfamily{pcr}\selectfont}

\lstdefinestyle{mystyle}{
    language=Python,
    basicstyle=\ttfamily\tiny,
    keywordstyle=\color{blue},
    commentstyle=\color{deepgreen},
    stringstyle=\color{deepred},
    breakatwhitespace=false,         
    breaklines=true,                 
    captionpos=b,                    
    keepspaces=true,                 
    showspaces=false,                
    showstringspaces=false,
    showtabs=false,                  
    tabsize=2,
    morekeywords={assert,True,False}
}

\title{Allegro: Open the Black Box of \\Commercial-Level Video Generation Model}

\author{Yuan Zhou, Qiuyue Wang, Yuxuan Cai, Huan Yang\thanks{corresponding author: huanyang@rhymes.ai} \AND Rhymes AI}

\begin{document}

\maketitle

\begin{figure}[h]
    \centering
    \vspace{-2ex}
    \includegraphics[width=\textwidth]{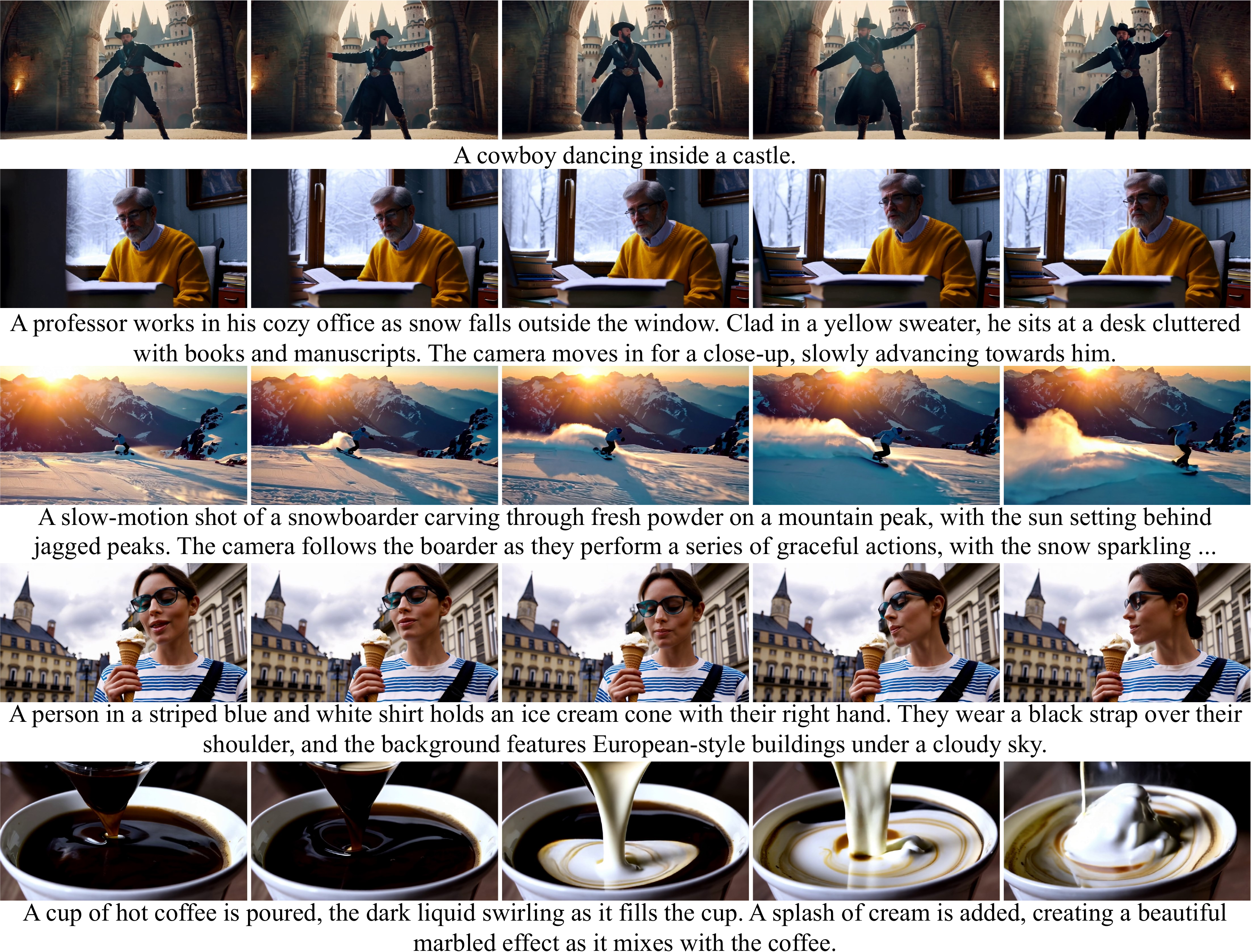}
    \caption{Allegro is capable of generating high-quality, dynamic videos from various descriptive text inputs.}
    \vspace{-2ex}
    \label{fig:teaser}
\end{figure}

\begin{abstract}
Significant advancements have been made in the field of video generation, with the open-source community contributing a wealth of research papers and tools for training high-quality models. However, despite these efforts, the available information and resources remain insufficient for achieving commercial-level performance. In this report, we open the black box and introduce \textbf{Allegro}, an advanced video generation model that excels in both quality and temporal consistency. We also highlight the current limitations in the field and present a comprehensive methodology for training high-performance, commercial-level video generation models, addressing key aspects such as data, model architecture, training pipeline, and evaluation. Our user study shows that Allegro surpasses existing open-source models and most commercial models, ranking just behind Hailuo and Kling.

\vspace{1ex}
Code: \url{https://github.com/rhymes-ai/Allegro}\\
Model: \url{https://huggingface.co/rhymes-ai/Allegro}\\
Gallery: \url{https://rhymes.ai/allegro_gallery}

\end{abstract}

\newpage

\section{Introduction}

The demand for video content in digital media has surged in recent years, driving innovation in automated video generation technologies. As these technologies evolve, the process of creating videos has become more accessible, efficient, and streamlined. Among these advancements, text-to-video generation models have emerged as a breakthrough. These models allow users to generate dynamic visual content from descriptive text, offering a highly flexible and controllable method for video creation. By transforming textual input into visual narratives, they are democratizing video production and opening new possibilities for content creators across industries.

With the advent of diffusion models, the field of visual content generation has experienced significant growth~\citep{ho2020denoising,ldm,saharia2022photorealistic,blattmann2023align,ho2022video,singer2022make,voleti2022mcvd,ruan2023mm,wang2024videocomposer,he2024dreamstory}. In the realm of image generation, open-source models have achieved performance levels comparable to many commercial products. Meanwhile, the field of text-to-video generation has seen the emergence of numerous commercial systems, such as Runway Gen-3~\citep{gen3}, Luma Dream Machine~\citep{lumalab}, OpenAI Sora~\citep{OpenAI2024b}, Kling~\citep{klingai}, Hailuo~\citep{hailuo}, and Pika~\citep{pikalab}.
However, unlike text-to-image generation, many key aspects of developing commercial-level text-to-video generation models are still unclear. These include efficiently modeling video data, accurately aligning videos with textual semantics, incorporating more precise controls, and developing high-quality text-video datasets for training.

In this report, we open the black box of commercial-level video generation models and introduce \textbf{Allegro}, an elite video generation model with exceptional quality and temporal consistency. We examine the critical elements that contribute to enhancing the model's output, offering detailed insights into the techniques and strategies employed for training a commercial-level video generation model. The report serves as an extensive guide for developing high-quality video generation models leveraging diffusion methods, with a focus on redesigning key components including:

\begin{itemize}
    \item \textbf{Data curation.} We propose a systematic data curation pipeline for training a commercial-level video generation model using large-scale image and video datasets, providing detailed insights into constructing optimized training data. With this pipeline, we build our dataset containing 106M images and 48M videos, with highly associated text captions.
    \item \textbf{Model architecture.} Building on the Diffusion framework, we introduce modifications to the Variational Autoencoder (VAE) and Diffusion Transformer (DiT) architectures to better accommodate the specific requirements of video generation. We outline strategies to optimize the computational infrastructure, ensuring efficient training and inference.
    \item \textbf{Evaluation.} We build a diverse benchmark for text-to-video generation and conduct user studies and subjective evaluations in six dimensions to ensure the generated videos meet aesthetic standards and align with human preferences.
\end{itemize}

Through the implementation of the aforementioned components, Allegro achieved results in subjective evaluations comparable to those of other commercial systems. As shown in Figure~\ref{fig:teaser}, Allegro is capable of generating high-quality, dynamic videos from various descriptive text inputs. Our user study demonstrates that Allegro outperforms current open-source models across all six dimensions and shows significant advantages over commercial models in most areas. Notably, it surpasses all commercial models in video-text relevance and ranks just behind Hailuo and Kling in overall quality.

In addition, we offer further insights and constructive guidance on how to enhance the base capabilities of the model, including model scaling-up, prompt refiner adaptation, and video tokenizer designing. Meanwhile, we are working on extending Allegro's foundational text-to-video model in two key directions: Image-to-Video with text conditions and more flexible motion control. These advancements aim not only to add more features to Allegro, but also to provide deeper insights on better aligning the model’s capabilities with user needs and boosting productivity. We release Allegro under the Apache 2.0 license, free for both academic and commercial use. 

\section{Data Curation}
Data curation is the primary task in building video generation models, permeating the entire training process.
Existing publicly available datasets, such as WebVid~\citep{bain2021frozen}, Panda-70M~\citep{chen2024panda}, HD-VILA~\citep{xue2022advancing}, HD-VG~\citep{wang2023videofactory} and OpenVid-1M~\citep{nan2024openvid}, have provided solid foundation for data sourcing and acquisition, offering diverse and extensive video data. However, with the sheer volume of data now available, significant challenges arise in terms of processing efficiency, data redundancy, and ensuring high-quality inputs for model training. 

One of the key issues with current data curation practices is the lack of efficient pipelines capable of handling and transforming these massive datasets into well-prepared, structured inputs. The time and computational resources required to process raw data are often prohibitive, leading to bottlenecks in the development cycle. Moreover, without carefully curated datasets, there is an increased risk of including noisy, redundant, or irrelevant data, which can degrade the performance of the models.

In addition to improving processing efficiency, another critical goal is to ensure that the curated datasets are not only large in volume but also diverse and of high quality. The success of video generation models hinges on the quantity of data and its richness in terms of variety, coherence, and prompt alignment. By addressing these issues, the objective is to build a robust data pipeline that optimizes the balance between data volume and quality, ultimately enhancing model performance in generating realistic and coherent videos.

In this section, we introduce a systematic and scalable data curation pipeline designed to meet the demands of large-scale video datasets. This pipeline aims to streamline the processing of raw video data and ensure that high-quality, diverse samples are used for training, providing insights for improving the overall dataset quality and efficiency.
We also provide a detailed analysis of the data distribution on different dimensions and hopefully give some insights on building a better training dataset.

\subsection{Data Processing}
The data processing pipeline plays a pivotal role in ensuring the quality and diversity of the data used for different stages of the model training process. 
This pipeline consists of three critical parts: data filtering, annotation, and stratification.

\subsubsection{Data Filtering}
\begin{figure}[t]
    \centering
    \includegraphics[width=0.9\textwidth]{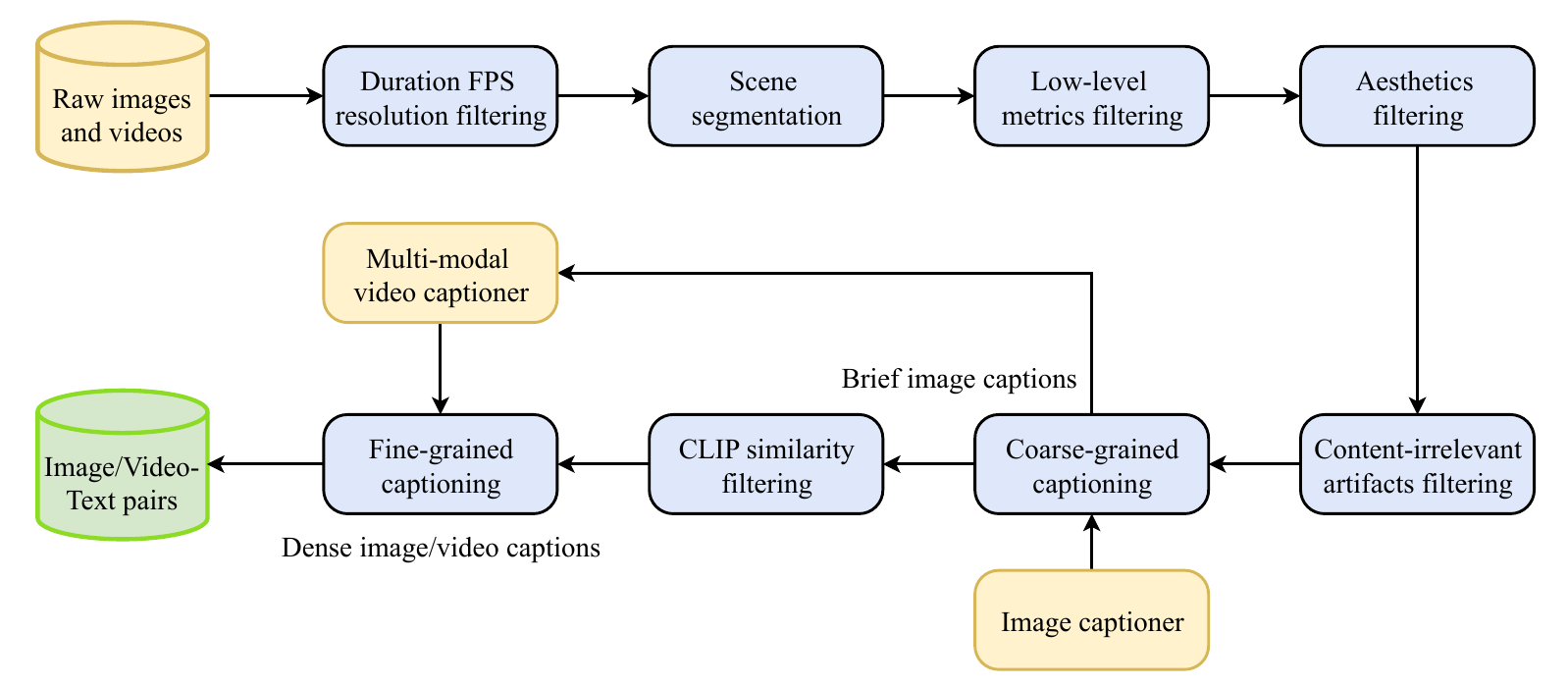}
    \caption{Allegro data filtering pipeline. A large pool of raw images and videos is fed into this pipeline to get a training set of high visual quality with associated text captions.}
    \label{fig:data_filtering}
\end{figure}

The data filtering pipeline is illustrated in Figure~\ref{fig:data_filtering}. Before the fine-grained video captioning, which will be introduced in the next section, we apply seven filtering steps in sequence: 
1) Duration and resolution filtering; 2) Scene segmentation; 3) Low-level metrics filtering; 4) Aesthetics filtering; 5) Content-irrelevant artifacts filtering; 6) Coarse-grained captioning; and 7) CLIP similarity filtering.
This pipeline is adaptable for both image and video data, with only minor adjustments for each data format. We introduce the details of each filtering step below. 
Due to the varying data quality requirements at different training stages, we focus on describing the specific methods of each data filtering step. 
The specific threshold values for certain dimensions will be detailed in Section~\ref{sec:data_strat}.

\textbf{Duration, FPS and Resolution Filtering.} 
We remove the raw images and videos with a resolution lower than 360p. We also remove raw videos that are shorter than 2 seconds or frame rates lower than 23 FPS. 
This step ensures that our dataset consists of high-quality image/video clips with sufficient length and visual fidelity for effective model training.

\textbf{Scene Segmentation.} 
We use PySceneDetect~\citep{PysceneDetect} to detect the scenes and divide the raw videos into single-scene video clips. 
We removed each video clip's first and last 10 frames to avoid false positives in scene detection. Only the clips with a duration of 2 to 16 seconds are kept to the next step.

\textbf{Low-level Metrics Filtering.} 
In this step, we focus on the visual quality of the data. We evaluate each video clip from four dimensions: brightness, clarity, semantic consistency, and motion amplitude.
We calculate the average grayscale of the images and the intermediate frame in video clips to filter out images and videos that are too dark or too bright.
The DOVER score~\citep{wu2023exploring} is used to assess images and video clips' clarity.
We employ LPIPS~\citep{zhang2018unreasonable} and UniMatch~\citep{xu2023unifying} to evaluate video clips' semantic consistency and motion amplitude, respectively. These two metrics are jointly considered to retain video clips with reasonable semantic consistency and motion amplitude.

\textbf{Aesthetics Filtering.}
We use the LAION Aesthetics Predictor~\citep{schuhmann2022laion} to evaluate the aesthetic score of images and the intermediate frames of each video clip, selecting the data with better visual appeal.

\textbf{Content-irrelevant Artifacts Filtering.}
The raw videos may contain artifacts unrelated to the video content, such as black borders, text, watermarks, etc., which can negatively affect video generation models. 
We use detection tools (e.g., the text detection tool CRAFT~\citep{baek2019character} and the watermark detection tool~\citep{watermark-detection}) to detect these elements and calculate their area ratio in the image or video frame. 
When the ratio is below the set threshold, we crop the original image or video; otherwise, the data will be discarded.

\textbf{Coarse-grained Captioning.}
We perform preliminary annotation on the remaining data to obtain coarse-grained captions, which provide global semantic information for images and videos. These coarse-grained captions also serve as input in the following \textbf{CLIP similarity filtering} and fine-grained captioning.
In this step, we use Tag2Text~\citep{huang2023tag2text} to caption the middle frames of the images and videos, which significantly improves the efficiency of obtaining coarse-grained descriptions during data processing.

\textbf{CLIP Similarity Filtering.}
The correlation of visual data and its caption is assessed by calculating the cosine similarity between the CLIP~\citep{radford2021learning} embeddings of the caption and those of the image or the video middle frame. We filter out samples with low CLIP cosine similarity to ensure more accurate inputs for subsequent fine-grained captioning and provide high-quality annotations for the final training data.

To facilitate the usability of the data, all video clips are uniformly re-encoded using H.264, and the frame rate is fixed at 30 FPS.
After the data filtering pipeline above, the final video clips are 2-16 seconds long, captured with a single-shot camera, and exhibit high visual quality, reasonable motion, and highly correlated coarse-grained captions.

\subsubsection{Data Annotation}
\label{sec:data_annotation}
\textbf{Fine-grained Captioning.}
We generate the fine-grained captions for both images and videos after data filtering by using Aria~\citep{li2024ariaopenmultimodalnative}, which has been finetuned on the video captioning task.
The coarse-grained captions are included in the video caption model's instructions to enhance annotation accuracy.
The fine-grained captions cover both the spatial and temporal information, including attributes of the subjects, interactions among the subjects, description of the backgrounds, environment, style, and atmosphere, camera angle and motion, and the process of the above content changing over time. To enhance the model's capability in camera control, some data includes video captions with sentences that begin with "Camera [MOTION\_PATTERN]." to explicitly highlight the camera motion.

Based on data annotations in both coarse- and fine-grained captioning stages, we obtained the final image/video-text data pairs. 
For image data, there are coarse- and fine-grained captions. For video data, there are coarse- and fine-grained captions for both the entire video and middle frames. More examples of the data captions are listed in Appendix \hyperref[sec:appx_data_analysis]{A}.

\subsubsection{Data Stratification}
\label{sec:data_strat}
The annotated data is organized into different datasets tailored to the needs of various training stages. 
Specifically, the data is stratified to support three main stages of progressive training: text-to-image pre-training, text-to-video pre-training, and text-to-video fine-tuning.
Table~\ref{tab:data_stratification} illustrates the specific threshold for different dimensions at different training stages.

\textbf{Text-to-Image Pre-training Data.} 
This dataset only contains the image-text pairs and is designed for training models to only generate images from textual descriptions. 
In this stage, the model learns how to map textual input to visual features (e.g., object shapes, textures, colors). It serves as a foundational step before moving on to video generation tasks. 
We finally get 107M images filtered from 412M raw images.

\textbf{Text-to-Video Pre-training Data.} 
This dataset contains video-text pairs that allow the model to learn how to generate coherent sequences of images (i.e., video) from text prompts.
The model builds upon the knowledge gained during text-to-image pre-training but is enforced to handle temporal consistency. 
This stage comprises two sub-stages based on different video resolutions (360p and 720p).
We keep 48M and 18M (out of 500M) video clips for the two sub-stages, respectively.

\textbf{Text-to-Video Fine-tuning Data.} 
This data is used in the final fine-tuning stage, where the model is optimized for video generation tasks. 
It ensures that the model produces high-quality, temporally consistent videos that align with the input text.
The model can better handle complex scenes, intricate motions, and detailed textures in the generated video.
Only around 2M video clips are selected for this final stage.

\begin{table}[t]
\centering
\resizebox{\textwidth}{!}{
\begin{tabular}{l |c|c|c|c}
\hline
\multirow{2}{*}{\makecell[c]{Filtering Step}} & \multirow{2}{*}{\makecell[c]{Text-to-Image \\ Pre-training}} & \multirow{2}{*}{\makecell[c]{Text-to-Video \\ Pre-training-360p}} & \multirow{2}{*}{\makecell[c]{Text-to-Video \\ Pre-training-720p}} & \multirow{2}{*}{\makecell[c]{Text-to-Video \\ Fine-tuning}} \\
&&&&\\
\hline
Duration & $-$ & $[2s, 16s]$ & $[2s, 16s]$, $[6s, 16s]$ & $[6s, 16s]$  \\
FPS & $-$ & $(23, 61)$ & $(23, 61)$ & $(23, 61)$  \\
Resolution (W=Width, H=Height) & W$\geq640$, H$\geq368$ & W$\geq640$, H$\geq368$ & W$\geq1280$, H$\geq720$ & W$\geq1280$, H$\geq720$ \\
Brightness & $[20, 180]$ & $[20, 180]$  & $[20, 180]$ & $[20, 180]$  \\
Clarity (DOVER Score) & $-$ & $-$ & $-$ & $\geq0.07$ \\
Semantic Consistency (LPIPS score) & $-$ & $-$ & $-$ & $\geq0.05$  \\
Motion (UniMatch Score) & $-$ & $-$ & $-$ & $[1.0, 100]$  \\
Aesthetics (Aesthetic Score) & $\geq4.8$ & $\geq4.8$ & $\geq5.0$ & $\geq5.3$ \\
Text Area Ratio in \% & $\leq0.05$ & $\leq0.05$ & $\leq0.05$ & $\leq0.05$  \\
CLIP Similarity & $\geq0.17$ & $\geq0.17$ & $\geq0.20$ & $\geq0.20$  \\
Remains/Total Data Counts & $107$M/$412$M & $48$M/$500$M & $18$M/$500$M & $2$M/$500$M  \\
\hline
\end{tabular}
}
\vspace{1ex}
\caption{Data filtering threshold across different training stages.}
\label{tab:data_stratification}
\vspace{-1ex}
\end{table}

\subsection{Data Analysis}
More details of data statistics, representative cases and data insights are in Appendix \hyperref[sec:appx_data_analysis]{A}.

\begin{figure}[t]
    \centering
    \includegraphics[width=0.95\textwidth]{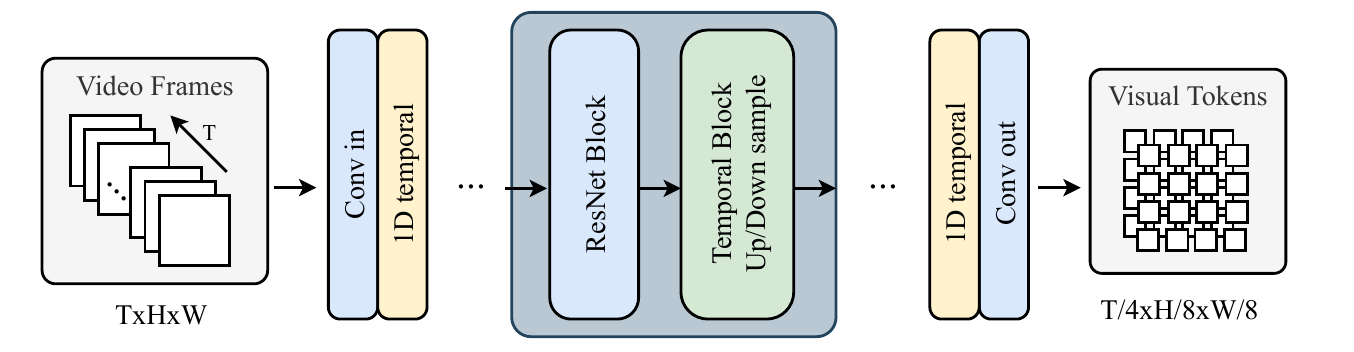}
    \caption{The architecture of our VideoVAE.
    The blue part represents the spatial layers of the original image VAE, while the other colored modules are the added temporal layers. With video frames of shape $T \times 3 \times H \times W$ as input to the VideoVAE encoder, we can obtain latent video representation with shape $T/S_{T} \times C_{l} \times H/S_{H} \times W/S_{W}$. The figure shows the model structure and process of the encoder part. The decoder part is symmetric and it takes the latent as input and reconstructs RGB videos.}
    \label{fig:vae}
\end{figure}

\section{Video Variational Auto-Encoder}
\label{sec:vvae}

Previous studies~\citep{ldm, SDXL} have demonstrated that the efficiency and performance of diffusion models can be significantly enhanced by adopting latent space modeling using a Variational Auto-Encoder (VAE)~\citep{kingma2013auto}.
The Sora~\citep{OpenAI2024b} team has also employed a video spatio-temporal compression network and trained Sora to generate videos within this compressed latent space. 
Consequently, a practical approach to generating long, high-resolution videos is to compress RGB videos in both spatial and temporal dimensions before modeling them in the latent space. 
The latent predicted by the Diffusion Transformer (DiT) is then decoded using a VAE decoder to produce the RGB videos. 
To facilitate this process, we constructed a Video Variational Auto-Encoder (VideoVAE), which encodes RGB videos into a compressed latent spatio-temporal space to improve the efficiency and performance of the diffusion model.

\subsection{Model Architecture}

The effectiveness of a VAE primarily depends on its ability to perform robust spatio-temporal modeling. To fully leverage the spatial compression capability of the existing image VAE, our VideoVAE design incorporates extended spatio-temporal modeling layers based on the image VAE. 
Specifically, We first adopt the architecture and weights of the image VAE from Playground v2.5~\citep{li2024playground} to model spatial information. 
Then, we incorporate temporal modeling layers to capture temporal information and apply temporal compression to reduce its dimensionality.

Our VideoVAE architecture is depicted in the Figure~\ref{fig:vae}. 
We add a 1D temporal CNN layer at the beginning and the end of both the encoder and the decoder in the image VAE.
Additionally, we design a temporal block consisting of four 3D CNN temporal layers, which are inserted after every ResNet block in VAE.
In temporal blocks, temporal downsampling is achieved through stridden convolution with a stride of 2, while upsampling is accomplished by frame repetition followed by a deconvolution operation.
After incorporating these temporal layers, the parameter count of our VideoVAE has reached 174.96M.

In the preliminary experiments, the temporal CNN layer within the temporal block had a kernel size of 1 in the spatial dimension (i.e., 1D temporal CNN layer). However, we discover that a spatial kernel size of 3 could accelerate convergence and achieve better reconstruction performance. Therefore, we employ 3D CNN layers in all upsampling and downsampling temporal blocks.

The reconstruction upper bound of VideoVAE is highly correlated with the spatial compression stride, temporal compression stride, and the number of dimensions in the latent space.
We continue to use the latent channel $ C_{l} = 4$ from image VAE without any expansion and have compression stride $S_{T}\times S_{H}\times S_{W} = 4\times 8\times 8$ in temporal, height and width dimension respectively. 
With video frames $T \times 3 \times H \times W $ as input to the VideoVAE encoder, we can obtain latent video representation with shape $T' \times 
C_{l} \times H' \times W' = T/S_{T} \times 
C_{l} \times H/S_{H} \times W/S_{W}$.

\subsection{Training Strategy}
The quality of reconstructed videos after compression and decompression by VideoVAE is closely tied to the quality of the training data. 
As a result, we prioritize videos and images with higher resolutions for training. Specifically, we only use images and videos where the shortest side exceeds 720 pixels. 
We ultimately select 54.7K videos and 3.73M images from our filtered dataset to train VideoVAE.

The effectiveness of VideoVAE in video reconstruction is strongly linked to the level of image detail. Since resizing images can reduce detail, our spatial data augmentation only includes random cropping. For temporal data augmentation in videos, we randomly choose a sampling interval from the set [1, 3, 5, 10] to sample frames evenly across video clips. Experiments demonstrate that using larger sampling intervals significantly accelerates the convergence of the temporal layers.

During model training, we exclusively utilize L1 loss and LPIPS loss~\citep{zhang2018unreasonable}, with respective loss weights of 1.0 and 0.1.
L1 loss, or mean absolute error (MAE), measures the average absolute differences between predicted and actual values.
LPIPS loss~\citep{zhang2018unreasonable} is a perceptual loss function that evaluates image similarity based on learned features from a deep neural network, aiming to capture human visual perception for tasks like image reconstruction and style transfer.

We conduct a two-stage training on VideoVAE. In the first stage, we perform joint training of the VideoVAE by incorporating both images and videos, utilizing a batch composition of one 16-frame video and four images, all with a spatial resolution of $256 \times 256$.
This stage is conducted for 65K iterations on 32 Nvidia A100 GPUs, with batch size 1 per GPU.

Image data will exclusively pass through the original image VAE's spatial layers, while video data will traverse both spatial and temporal layers. We anticipate that this mixed training strategy will not only effectively enhance the model's spatiotemporal modeling capabilities but also maintain its excellent spatial modeling capabilities through the utilization of images.

In the second stage, we freeze the spatial layers from the image VAE and fine-tune the temporal layers using 24-frame $256 \times 256$ resolution videos to enhance temporal modeling capabilities. In this stage, we fine-tune the temporal layers for 25K iterations on 32 Nvidia A100 GPUs, with batch size 1 per GPU.

\subsection{Inference Strategy}
Encoding and decoding high-resolution, long-duration videos directly is impractical due to memory limitations. To handle such videos during inference, we divide both the input video and the latent tensor into spatio-temporal tiles. Each tile is then encoded and decoded separately, and the results are reassembled to form the final output.
In practice, we employ a tile size of $24 \times p \times p$ (time, height, width) in Encoder, and a tile size of $6 \times p/8 \times p/8$ in Decoder.
While blending and reassembling the output, we choose an overlapping stride $ v_{T} \times v_{H} \times v_{W}$ among pixel tiles and $ v_{T}/4 \times v_{H}/8 \times v_{W}/8$ among latent tiles.
For 320p video generation, we take $ p=256,  v_{T} \times v_{H} \times v_{W}=8\times 64\times 144$. And we have $ p=320, v_{T} \times v_{H} \times v_{W}=8\times 80\times 120$ for 720p video generation.
Following the approach of the image VAE, we utilize a linear combination in both spatial and temporal dimensions for blending.
For image encoding and decoding, we only use the spatial layers in VideoVAE, which is consistent with those in the training process.

\section{Video Diffusion Transformer}
The Video Diffusion Transformer (VideoDiT) represents a breakthrough in video generation by combining the strengths of diffusion models with the Transformer architecture~\citep{ho2022video, ma2024latte, dinesh2022walt, yang2024cogvideox, lu2023vdt, chen2024gentron}. VideoDiT excels at modeling both spatial and temporal dependencies, ensuring improved coherence across video frames. This approach generates high-quality, realistic videos while scaling efficiently on larger datasets, outperforming traditional methods. Several benchmarks consistently demonstrate DiT’s superior performance in video generation tasks~\citep{huang2024vbench, liu2024evalcrafter, sun2024t2v, liu2024fetv}.

Recently, open-source projects such as Open-Sora~\citep{opensora} and Open-Sora-Plan~\citep{OpenSORAPlan2024} have significantly advanced the field of video generation. Notably, the Open-Sora-Plan has been widely adopted for its robust implementations. Our work builds upon the VideoDiT architecture introduced in Open-Sora-Plan v1.2.0 and aims at opening the black box of training a commercial-level video generation model.

We introduce these key modifications to improve performance. First, we replace the original mT5 text encoder~\citep{xue2020mt5} with T5~\citep{raffel2020exploring}, enhancing text-to-video alignment and improving both quality and coherence. Second, we replace the causal VideoVAE with a custom-trained VideoVAE to address issues of temporal consistency and frame quality. Finally, we train the modified framework on a large-scale curated dataset with a multi-stage training strategy, achieving a commercial-level video generation model.

\subsection{Model Architecture}
\label{sec:dit_arch}
\begin{figure}[t]
    \centering
    \includegraphics[width=0.9\textwidth]{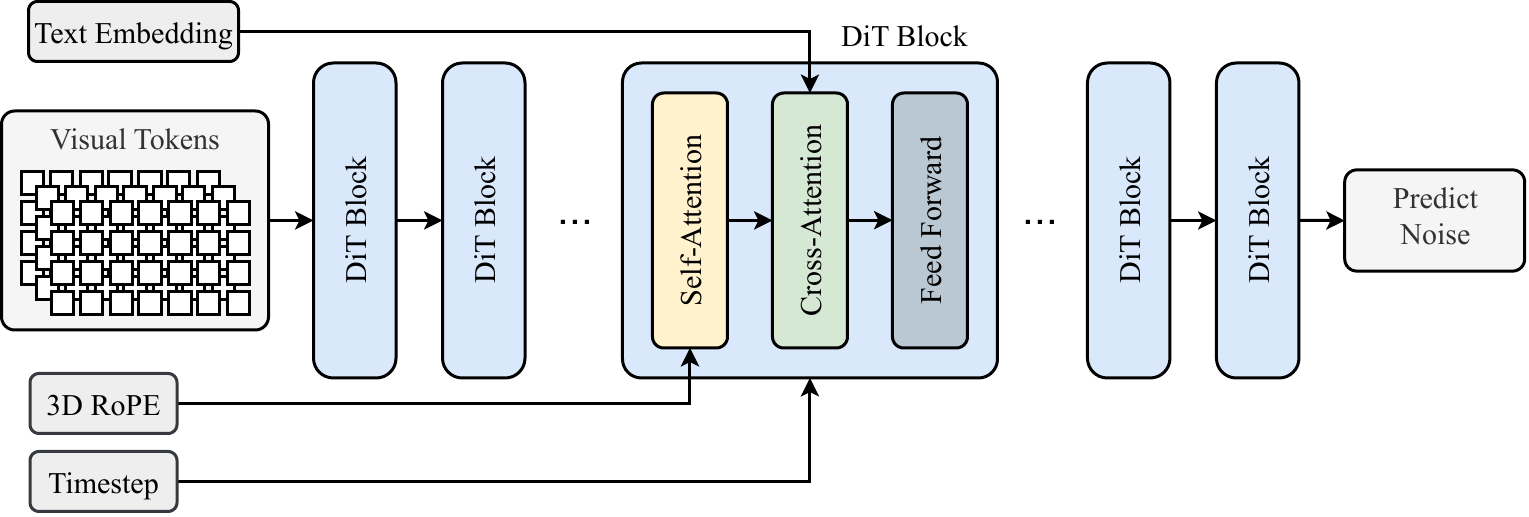}
    \caption{The architecture of our VideoDiT. The T5 text encoder converts natural language into text embeddings. The visual token is encoded by our proposed VideoVAE encoder.
    The video Transformer network, composed of multiple DiT blocks, predicts the noise based on visual tokens, text embedding, diffusion timestep, and 3D RoPE.}
    \label{fig:videodit}
\end{figure}

As shown in Figure~\ref{fig:videodit}, a typical VideoDiT framework comprises three key modules that collaboratively generate videos from textual descriptions: 1) a text encoder that converts natural language into text embeddings to guide video generation; 2) a VideoVAE that encodes videos into a latent space for efficient diffusion modeling; and 3) a video Transformer network that processes both visual tokens and text embeddings to predict noise for the diffusion generation.

\textbf{Text-Encoder.}
The text encoder is essential in video diffusion models, translating natural languages into meaningful embeddings that guide video generation. Early models like Stable Diffusion~\citep{ldm, SDXL} adopted the CLIP text encoder~\citep{radford2021learning}, effectively aligning text with visual features. However, as tasks have become more complex, the T5 text encoder has gained popularity for its stronger language understanding and ability to produce richer embeddings, leading to better video quality and coherence. In Open-Sora-Plan v1.2.0, the mT5 text encoder offers improved multilingual compatibility, but the predominance of English captions in the training data limits the model's efficiency in aligning textual semantics and visual information. To address these limitations, we adopt T5 as our text encoder based on prior visual generation research~\citep{chen2024gentron, saharia2022photorealistic, DeepFloydIF}. Experiments in Section~\ref{sec:eval_videodit} show that this choice significantly enhances the model's semantic coherence, resulting in better text-to-video modeling. By leveraging T5, we achieve a more effective integration of textual and visual features, ultimately improving the quality and relevance of the generated videos.

\textbf{VideoVAE.}
The VideoVAE module, crucial for compressing high-dimensional video data, is frequently employed in video diffusion models to improve efficiency.
The idea of introducing VAEs into diffusion models was first proposed in Latent Diffusion Models (LDM)~\citep{ldm}, to enable effective training of text-to-image generation tasks. This integration has been expanded in subsequent efforts like Stable Diffusion~\citep{esser2024scaling, SDXL}.
Given the greater computational demands and information density of video data, we employ a VAE to compress videos before the VideoDiT module, significantly enhancing training efficiency and convergence speed. As discussed in Section~\ref{sec:vvae}, we have developed a VideoVAE that outperforms existing open-source alternatives, offering superior visual quality and smoother motion.
Our advanced VideoVAE can effectively encode videos, achieving a high compression ratio of $4\times8\times8$, while simultaneously preserving both motion dynamics and static details.

\textbf{Video Transformer.}
The core of VideoDiT comprises a Video Transformer built from several Diffusion Transformer blocks (DiT blocks~\citep{DiT, chen2023pixart}). Each DiT block includes these modules: 1) a self-attention module with 3D RoPE that models video tokens across both spatial and temporal dimensions; 2) a cross-attention module that injects textual conditions to guide the generation process; 3) a feed-forward layer; and 4) a set of AdaLN-single modules~\citep{perez2018film} that incorporate information from the diffusion timestep. The self-attention module is particularly critical for the quality of the generated videos. Earlier works tend to employ a combination of 2D spatial and 1D temporal attention to approximate full 3D attention. However, this limitation often results in videos with poor continuity and minimal motion. Recent research indicates that 3D attention significantly enhances video generation quality~\citep{OpenSORAPlan2024, polyak2024moviegencastmedia}. Adopting this strategy, we generate videos with improved consistency and dynamic motion.

More details about the model specifications can be found in Appendix~\ref{sec:appx_modelspec}.

\subsection{Multi-Stage Training Strategy}
\label{sec:dit_training}
\begin{table}
\centering
\resizebox{\textwidth}{!}{
\begin{tabular}{c|c|ccc|c}
\hline
 Param. & T2I Pre-train & \multicolumn{3}{c|}{T2V Pre-train} & T2V Fine-tune \\ \hline
 Target & 1$\times$368$\times$640 & 40$\times$368$\times$640 & 40$\times$720$\times$1280 & 88$\times$720$\times$1280 & 88$\times$720$\times$1280 \\
 GPUs & 128$\times$H100 & 256$\times$H100 & 256$\times$H100 & 256$\times$H100 & 256$\times$H100 \\
 Batch-size & 4,096 & 1,024 & 512 & 256 & 256 \\
 Steps & 170K & 85K & 41K & 31K & 10K \\
 Data & 700M & 87M & 21M & 8M & 2.6M \\   
 \hline
\end{tabular}
}
\vspace{1ex}
\caption{Training details and resource requirements for different stages.}
\label{tab:dit_training}
\end{table}

Our model training process can be divided into three stages: 1) the text-to-image pre-training stage, which establishes a mapping between text and images; 2) the text-to-video pre-training stage, which builds on this mapping to learn the motion relationships between objects; and 3) the text-to-video fine-tuning stage, which focuses on improving the overall visual quality of the generated videos. More details are listed in Table~\ref{tab:dit_training}.

\textbf{Text-to-Image Pre-training.}
The text-to-image stage aims to establish the connection between textual descriptions and visual elements. To improve training efficiency, we begin with the pre-trained text-to-image model from Open-Sora-Plan v1.2.0 and adjust the target image resolution to $368\times 640$. We replace the text encoder with T5 (512 max token length) and the VAE with our custom-trained VideoVAE at this stage, as discussed in Section~\ref{sec:dit_arch}. 
To better align with the subsequent text-to-video training, we incorporate static video frames as images for mixed training. This stage is trained using 128 Nvidia H100 GPUs with a batch size of 4,096, completing 170K steps and processing a total of 700M images.

\textbf{Text-to-Video Pre-training.}
We employ a three-stage progressive strategy to pre-train our text-to-video model. This pre-training is extended from our previous text-to-image model to learn dynamics from paired text-video data.
First, we train with 40 frames at $368\times640$ resolution to reduce the computational cost, utilizing 256 Nvidia H100 GPUs. This sub-stage has trained for 85K steps with a batch size of 1,024, processing a total of 87M videos, to equip the model with basic dynamics generation ability.
Next, we scale up the resolution to $720\times1280$ while keeping the video length. We resume from the previous sub-stage and train for 41K steps over 21M videos, using 256 Nvidia H100 GPUs with a batch size of 512 for 41K steps. After this stage, the model's capabilities improved further, with the ability to generate higher-quality videos. However, compared to the detailed textual descriptions in the training data, the relatively short video duration can lead to some distortion and motion warping.
In the final sub-stage, we increase the video length to 88 frames at $720\times1280$ resolution. We use 256 Nvidia H100 GPUs with a batch size of 256, training for 31K steps and processing a total of 8M videos. All videos across the three sub-stages are sampled at 15 FPS, enabling the model to generate 6-second videos at $720\times1280$ resolution.

\textbf{Text-to-Video Fine-tuning.}
To further enhance video generation quality, we select a batch of high-quality dynamic videos and mix in various types of video captions for further fine-tuning. In this stage, we resume from the pre-trained model and train for 10K steps with 2.6M videos in total, using 256 Nvidia H100 GPUs with a batch size of 256. 
This fine-tuning enables the model to handle a wider variety of text inputs, both in length and format, and generate high-quality dynamic videos.

\subsection{Infrastructure Optimzation}
We train our VideoDiT model using up to 256 Nvidia H100 GPUs with several optimization techniques. 

\textbf{Decoupled Inference of VAE.} 
We observe that the VideoVAE processing and text encoding account for over $30\%$ of the training computations. Since the inference process demands less VRAM and computation, we move the VideoVAE, text encoder to the mid-end inference GPUs, reserving the H100's high performance and large VRAM for longer contexts and larger batch sizes in training.
By JIT-compiling PyTorch code into optimized kernels with \textit{torch.compile} and batching inference, we reduce the inference time of VideoVAE and text encoding during training by $62\%$. To avoid redundant computations, we cache the inference results offline for the high-quality data trained over multiple epochs.

\textbf{Efficient Attention.}
We employ full bi-directional attention in the VideoDiT. For long sequences, attention accounts for $63.3\%$ of computations during training and $84.5\%$ during inference.
We utilize FlashAttention2~\citep{dao2023flashattention} in self-attention and XFormers memory-efficient attention~\citep{xFormers2022} (support mask) in cross-attention. Switching from XFormers to FlashAttention2 accelerates training in self-attention by $22\%$.

\textbf{Context Parallelism.}
Our model supports 88 frames at $720\times1280$ resolution video, which requires \textasciitilde 79.2K context length during training. To support longer context, we include Context Parallel (CP) in training. Instead of having a single device process the entire input, the input context is divided into smaller chunks, each processed in parallel by different devices. In \textit{softmax-qk-attention} forward and backward, each device conducts necessary communications with other devices, ensuring that all devices have the full activations or gradients during computation. Ring-Attention~\citep{liu2023ring} and Deepspeed-Ulysses~\citep{jacobs2023deepspeed} provide efficient implementation. We also include CP in inference time.

\section{Model Evaluation}
We evaluate the VideoVAE and VideoDiT of our proposed Allegro model in Section~\ref{sec:eval_VideoVAE} and Section~\ref{sec:eval_videodit}, respectively.

\subsection{Video Variational Auto-Encoder}
\label{sec:eval_VideoVAE}
Our VideoVAE and existing open-source 3D VAEs (such as Open-Sora v1.2~\citep{opensora}, Open-Sora-Plan v1.2.0~\citep{OpenSORAPlan2024}) share the same latent channel $C_{l} = 4$, and employ identical compression strides of $S_{T}\times S_{H}\times S_{W} = 4\times 8\times 8$ in the temporal, height, and width dimensions, respectively. 
We assess the performance of all available open-source VideoVAEs on a validation set comprising 100 videos, each with 120 frames and 720p resolution, utilizing metrics such as SSIM (Structural Similarity Index)~\citep{wang2004image}, PSNR (Peak Signal-to-Noise Ratio)~\citep{wang2004image}, and subjective evaluation criteria.
PSNR~\citep{wang2004image} is an image quality evaluation metric based on the Mean Squared Error (MSE). It assesses the quality of an image by comparing the differences between the original image and the processed image.
SSIM~\citep{wang2004image} is a quality evaluation metric based on structural similarity in images. It not only considers pixel-level differences but also considers the luminance, contrast, and structural information of the image.
For subjective evaluation criteria, we focus on identifying flickering and semantic distortions of videos that are not acceptable to human perception.

We observe that our VideoVAE model is highly sensitive to the temporal dimension of the tiles, with the best video reconstruction achieved when the tile duration matches the duration used during training. Deviations from the training duration lead to unpredictable flickering in the reconstructed videos. While we experiment with training on videos of varying lengths, this does not significantly enhance VideoVAE's ability to model videos of different durations effectively. Fixing the video clip length to match the training duration yields the best performance during inference. Our analysis shows that when the clip length is too short, the model becomes overly sensitive, resulting in suboptimal performance.

\begin{figure}[t]
    \centering
    \includegraphics[width=0.9\textwidth]{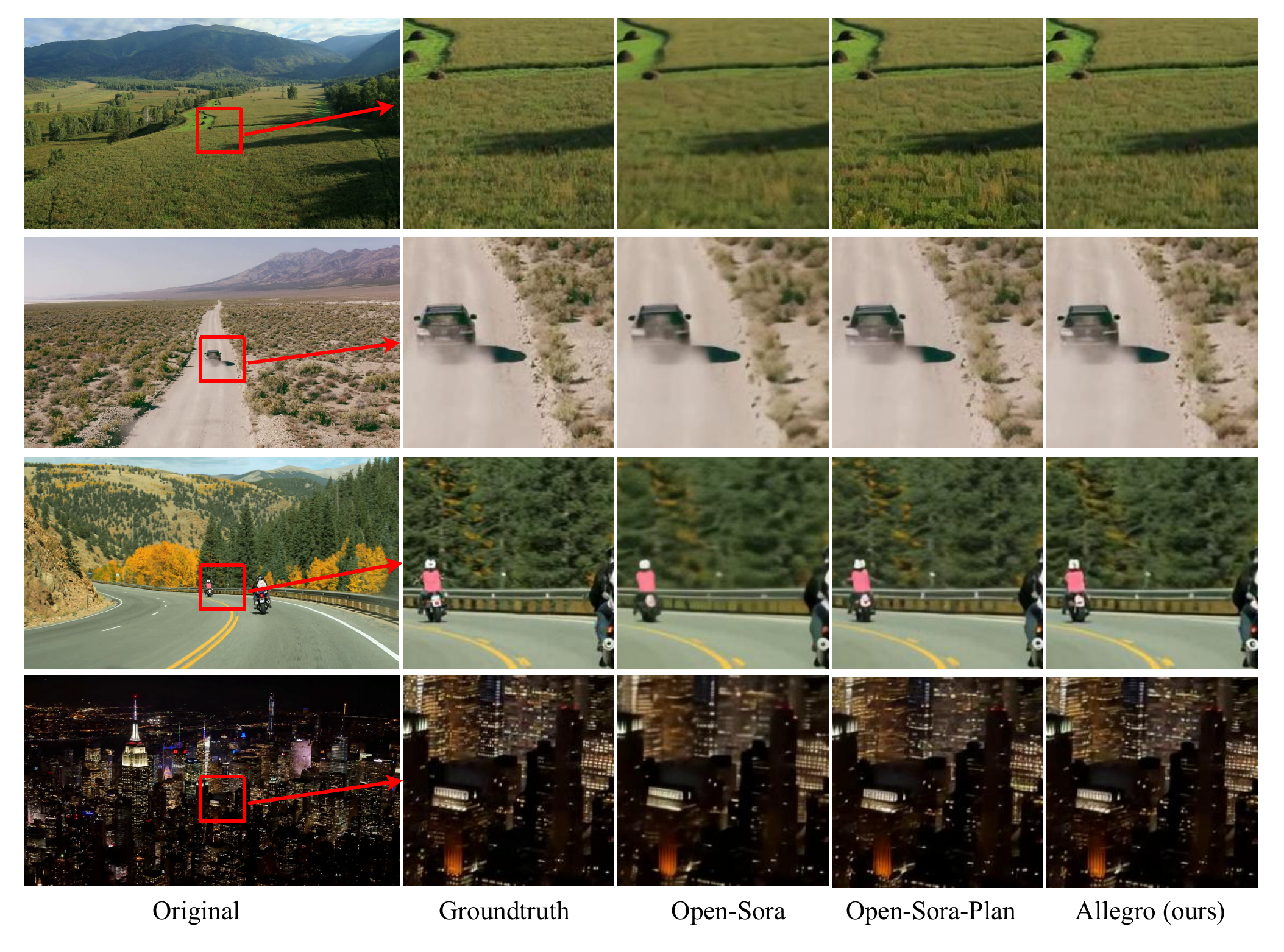}
    \caption{Visualization of Zoom-In on the sampled frame from reconstructed videos of VideoVAEs in our validation set.}
    \label{fig:vae_eval}
\end{figure}

\begin{table}[t]
    \centering
    \begin{tabular}{p{6cm}<{\centering} p{2cm}<{\centering} p{2cm}<{\centering}}
    \hline
    Method & PSNR$\uparrow$ & SSIM$\uparrow$ \\
    \hline
    Open-Sora & 30.64 & 0.8338 \\
    Open-Sora-Plan & 27.65 & 0.8173 \\
    \hline
    \textbf{Allegro (ours)} & \textbf{31.25} & \textbf{0.8553} \\
    \hline
    \end{tabular}
    \vspace{1ex}
    \caption{Comparison of PSNR and SSIM among current open-source VideoVAE models.}
    \label{tab:vae}
\end{table}

As shown in Table~\ref{tab:vae}, our VideoVAE outperforms other open-source models in both PSNR~\citep{wang2004image} and SSIM~\citep{wang2004image}, achieving values of 31.25 and 0.8553, respectively. 
In subjective evaluations, the reconstructions from our model show reduced flickering and avoid issues of excessive sharpening or over-smoothing. 
The first column in Figure 5 displays the original frame, while the second column shows a zoomed-in view of the original frame. 
The third, fourth, and fifth columns provide zoomed-in visualizations using Open-Sora~\citep{opensora}, Open-Sora-Plan~\citep{OpenSORAPlan2024}, and our VideoVAE, respectively. 
In the first row, the lawn reconstruction demonstrates that our VideoVAE avoids both excessive sharpening and blurring, compared to other methods. 
In the third row, our model effectively preserves the shape and details of the helmet, while in the fourth row, it better retains the details of the skyscraper lights than the other methods.

\subsection{Video Diffusion Transformer}
\label{sec:eval_videodit}
We evaluate the performance of our proposed Allegro model in this section. As our model outputs videos with 88 frames in 15 FPS and 720p resolution, we adopt a video frame interpolation model~\citep{zhang2023extracting} to enhance the FPS to 30 for better visual quality. 

\begin{table}[t]
    \centering
    \small
\begingroup
\renewcommand{\arraystretch}{1} 
    \setlength{\tabcolsep}{4pt}
    \begin{tabular}{c|ccc|cccccc}
    \hline
    Model & \makecell{Total \\ Score}  & \makecell{Quality \\ Score} & \makecell{Semantic\\ Score}  & \makecell{Subject \\ Consist.} &  \makecell{Background \\ Consist.} & \makecell{Aesthetic \\  Quality}  &  \makecell{Human \\  Action} & Scene \\
    \hline
    Open-Sora-Plan & 78.00 & 80.91 & 66.38 & 95.73 & 96.73 & 56.85 & 86.8 & 27.17 \\
   Open-Sora & 79.76 & 81.35 & 73.39 & 96.75 & 97.61 & 56.85 & 91.2 & 42.44 \\
   Pika & 80.69 & 82.92 & 71.77 & 96.94 & 97.36 & 62.04 & 86.2 & 49.83 \\
    CogVideoX & 80.91 & 82.18 & 75.83 & 96.78 & 96.63 & 60.82 & 98.0 & 51.14 \\
    Kling & 81.85 & 83.39 & 75.68 & 98.33 & 97.60  & 61.21 & 93.4 & 50.86 \\
    Gen-3 & 82.32 & 84.11 & 75.17 & 97.10 & 96.62 & 63.34 & 96.4 & 54.57 \\
    Hailuo & 83.41 & 84.85 & 77.65 & 97.51 & 97.05  & 63.03 & 92.4 & 50.68 \\
    \hline
    \textbf{Allegro (ours)} & \textbf{81.09} & \textbf{83.12} & \textbf{72.98} & \textbf{96.33} & \textbf{96.74} & \textbf{63.74} & \textbf{91.4} & \textbf{46.72} \\
    \hline
\end{tabular}
\endgroup
\vspace{1ex}
    \caption{Comparison of Allegro and other video generation models on VBench. The results are presented as percentages, with higher values indicating better performance. The version and date of each model can be found in the main text.}
    \label{tab:eval_vbench}
\end{table}

\textbf{Quantitative Evaluation.}
We use VBench~\citep{huang2024vbench} for our quantitative evaluation, running our model on the 946 text prompts provided by VBench. To reduce randomness, we generate five videos for each prompt, following VBench's guidelines. For prompts containing fewer than 9 words, we utilize a large language model (LLM) to refine them, enhancing robustness. In total, we generated 4,730 videos for the evaluation. Given the large number of videos required by VBench, we directly use the results from their leaderboard for other models.

As shown in Table~\ref{tab:eval_vbench}, we compare our model with Open-Sora v1.2 (8s), Open-Sora-Plan v1.1.0, CogVideoX-2B, Pika 1.0 (Jun. 2024), Gen-3 (Jul. 2024), Hailuo (MiniMax-Video-01) and Kling (Jul. 2024, high-performance mode). The results show that our method outperforms other open-source methods and ranks just behind Gen-3, Kling, and Hailuo among commercial models. For a detailed evaluation of all dimensions in VBench, please refer to the Appendix~\ref{sec:appx_vbench}.

\textbf{Qualitative Evaluation.}
Evaluating video generation results is a highly subjective task, and traditional quantitative metrics often show a significant deviation from the subjective experiences of human users. Therefore, we further conduct a user study to assess the subjective experience of the models.

To ensure the accuracy of the evaluation, we collect 46 diverse text inputs. These texts are from different sources: 1) text prompts from open-source models; 2) text prompts from commercial models; 3) captions generated by video captioning models; 4) texts constructed by large language models (LLMs) for video generation; and 5) real user inputs. The scenarios described in the texts cover a wide range, including people, animals, food, vehicles, landscapes, street scenes, sci-fi, and multi-object scenarios. The text lengths varied from five words to 96 words.

We compare our model with open-sourced models including Open-Sora-Plan v1.2.0, Open-Sora v1.2, CogVideoX-2B and commercial models including PixVerse v2 (Aug. 2024), Pika 1.0 (Aug. 2024), DreamMachine v1.6 (Aug. 2024), Vidu (Sept. 2024), Gen-3 (Sept. 2024), Hailuo (Sept. 2024) and Kling v1.5 (Sept. 2024). To simplify the naming, we will no longer include dates or version numbers in the names in the following sections.

To minimize ambiguity and obtain relatively accurate labeling results, we design six evaluation dimensions for annotators: 1) video-text relevance; 2) appearance distortion; 3) appearance aesthetics; 4) motion naturalness; 5) motion amplitude; and 6) overall quality. The evaluation is conducted using an A-B test format, where subjects are asked to rate whether A is better, B is better, or whether A and B are indistinguishable. For each trial, we randomly present results from our method and other methods, along with the corresponding text, and ask the annotators to rate them. Each video pair is rated by two subjects, and we collect a total of 5,448 ratings.

\begin{table}[t]
    \centering
    \begin{tabular}{l|ccc|ccc|ccc}
    \hline
    Dimensions & \multicolumn{3}{c|}{V.-T. Relevance} & \multicolumn{3}{c|}{Appearance Dist.} & \multicolumn{3}{c}{Appearance Aest.} \\
    \hline
    Methods & Win & Lose & Tie & Win & Lose & Tie & Win & Lose & Tie \\
    \hline
    Open-Sora-Plan & $\textbf{96\%}$& $0\%$& $4\%$& $\textbf{57\%}$& $11\%$& $32\%$& $\textbf{89\%}$& $0\%$& $11\%$\\
    Open-Sora & $\textbf{78\%}$& $9\%$& $13\%$& $\textbf{87\%}$& $4\%$& $9\%$& $\textbf{96\%}$& $4\%$& $0\%$\\
    CogVideoX & $\textbf{83\%}$ & $4\%$ & $13\%$ & $\textbf{67\%}$& $7\%$ & $26\%$ & $\textbf{96\%}$ & $4\%$ & $0\%$ \\
    \hline

    PixVerse & $\textbf{57\%}$& $2\%$& $41\%$& $\textbf{46\%}$& $17\%$& $37\%$& $\textbf{61\%}$&$11\%$ & $28\%$\\
    Pika & $\textbf{72\%}$& $2\%$& $26\%$& $\textbf{43\%}$& $20\%$& $37\%$& $\textbf{72\%}$& $9\%$& $20\%$\\
    Dream Machine & $\textbf{33\%}$& $22\%$& $45\%$& $\textbf{39\%}$& $13\%$& $48\%$& $17\%$& $35\%$& $48\%$\\
    Vidu & $\textbf{27\%}$& $15\%$& $58\%$& $\textbf{27\%}$& $22\%$& $51\%$& $\textbf{51\%}$& $29\%$& $20\%$\\
    Gen-3 & $\textbf{42\%}$& $18\%$& $40\%$& $\textbf{29\%}$& $20\%$& $51\%$& $\textbf{45\%}$& $33\%$& $22\%$\\
    Hailuo & $\textbf{33\%}$& $25\%$& $42\%$& $14\%$& $26\%$& $60\%$& $26\%$& $56\%$& $18\%$\\
    Kling & $\textbf{29\%}$& $27\%$& $44\%$& $4\%$& $36\%$& $60\%$& $5\%$& $82\%$& $13\%$\\
    
    \hline
    \hline
    Dimensions & \multicolumn{3}{c|}{Motion Nat.} & \multicolumn{3}{c|}{Motion Amp.} & \multicolumn{3}{c}{Overall Quality} \\
    \hline
    Methods & Win & Lose & Tie & Win & Lose & Tie & Win & Lose & Tie \\
    \hline
    Open-Sora-Plan & $\textbf{44\%}$& $17\%$& $39\%$& $\textbf{98\%}$& $2\%$& $0\%$& $\textbf{100\%}$& $0\%$& $0\%$\\
    Open-Sora & $\textbf{94\%}$& $2\%$& $4\%$& $\textbf{65\%}$& $33\%$& $2\%$& $\textbf{98\%}$& $2\%$& $0\%$\\
    CogVideoX & $\textbf{76\%}$& $9\%$& $15\%$& $\textbf{74\%}$& $24\%$& $2\%$& $\textbf{96\%}$& $4\%$& $0\%$\\
    \hline
    PixVerse & $\textbf{50\%}$& $13\%$& $37\%$& $\textbf{76\%}$& $9\%$& $15\%$& $\textbf{96\%}$& $4\%$& $0\%$\\
    Pika & $\textbf{43\%}$& $17\%$& $39\%$& $\textbf{91\%}$& $4\%$& $4\%$& $\textbf{89\%}$& $11\%$& $0\%$\\
    Dream Machine & $\textbf{28\%}$& $13\%$& $59\%$& $41\%$& $48\%$& $11\%$& $\textbf{61\%}$& $28\%$& $11\%$\\
    Vidu & $\textbf{40\%}$& $33\%$& $27\%$& $18\%$& $60\%$& $22\%$& $\textbf{51\%}$& $42\%$& $7\%$\\
    Gen-3 & $\textbf{24\%}$& $20\%$& $56\%$& $38\%$& $58\%$& $4\%$& $\textbf{49\%}$& $40\%$& $11\%$\\
    Hailuo & $14\%$& $21\%$& $65\%$& $35\%$& $46\%$& $19\%$& $47\%$& $49\%$& $4\%$\\
    Kling & $13\%$& $36\%$& $51\%$& $13\%$& $76\%$& $11\%$& $22\%$& $76\%$& $2\%$\\
    
    \hline
    \end{tabular}
    \vspace{1ex}
    \caption{User study on our benchmark with 46 text prompts. The performance is assessed in six dimensions: video-text relevance, appearance distortion, appearance aesthetics, motion naturalness, motion amplitude, and overall quality. The version and date of each model can be found in the main text. The winning dimension is highlighted in \textbf{bold}.}
    \label{tab:eval_userstudy}
\end{table}

As shown in Table~\ref{tab:eval_userstudy}, our model surpasses the current open-source models across all six dimensions. Compared to commercial models, our model demonstrates significant advantages in most aspects. 
Notably, in terms of video-text relevance, our model outperforms all commercial models. For the dimension of overall quality, our model outperforms most commercial models, ranking just behind Hailuo and Kling. 
However, the result also shows that there is still room for improvement in our model's handling of large-scale motion compared to the top-performing commercial models. This can typically be addressed by increasing the amount of accelerated motion in the training data or further filtering the speed of the videos used for training. Increasing the model's parameter size may also help improve its ability to model large-scale motion.

\begin{figure}[t]
    \centering
    \includegraphics[width=\textwidth]{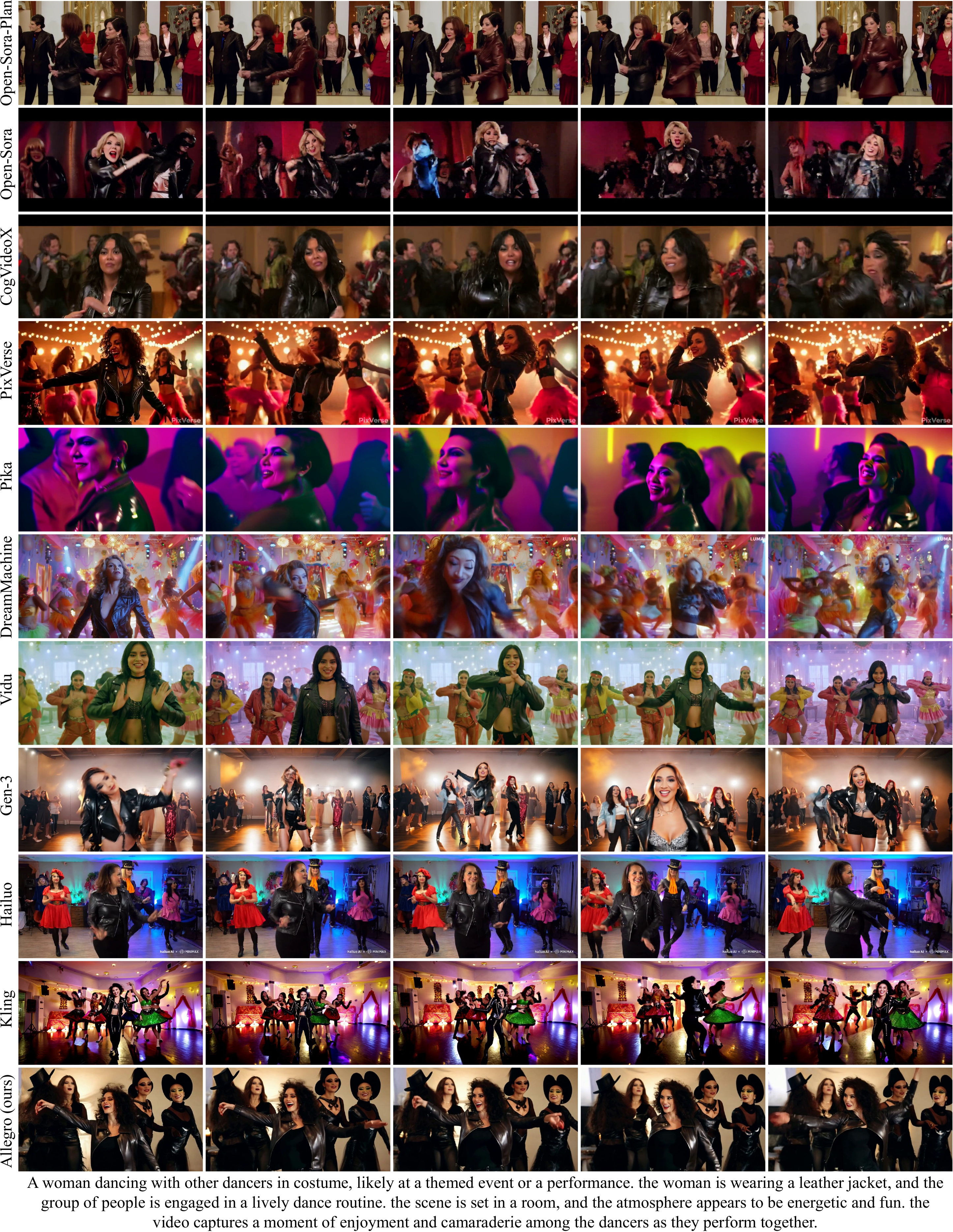}
    \caption{Qualitative comparisons of our Allegro with SOTA methods for the user study.}
    \label{fig:eval_userstudy}
\end{figure}

We show the visual result of our method and other methods in Figure~\ref{fig:eval_userstudy}. We can see that our model surpasses other models in both video-text relevance, video appearance quality, and video motion naturalness. More comparison results can be found in Appendix~\ref{sec:appx_morecmp}.

\begin{figure}[t]
    \centering
    \includegraphics[width=\textwidth]{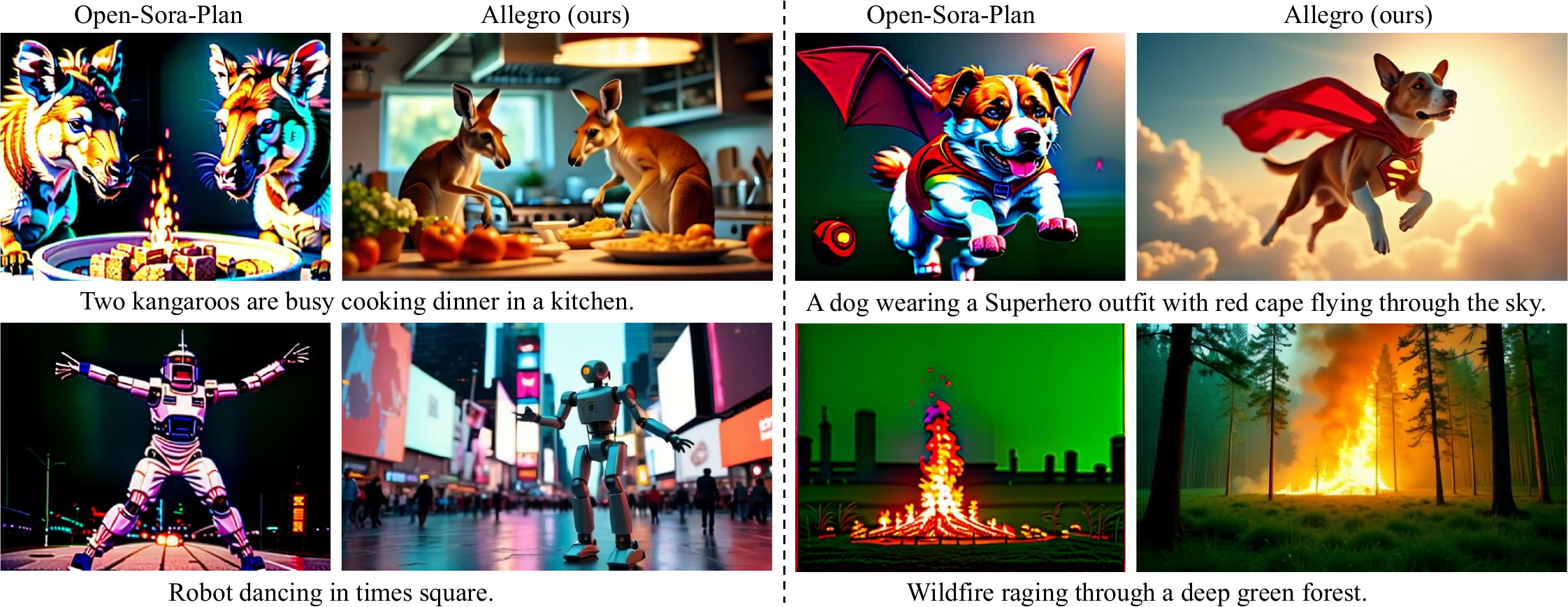}
    \caption{Qualitative comparison of generated images from our Allegro and the Open-Sora-Plan model in text-to-image generation. }
    \label{fig:eval_textencoder_t2i}
    \vspace{-1ex}
\end{figure}

\textbf{Text-Encoder.}
As discussed in Section~\ref{sec:dit_arch}, we compare our fine-tuned text-to-image model with T5 text-encoder with the original text-to-image model provided in Open-Sora-Plan that uses mT5 text encoder. As shown in Figure~\ref{fig:eval_textencoder_t2i}, images generated with Open-Sora-Plan lack visual details and have less relevance to the given text inputs. 

\begin{figure}[t]
    \centering
    \includegraphics[width=\textwidth]{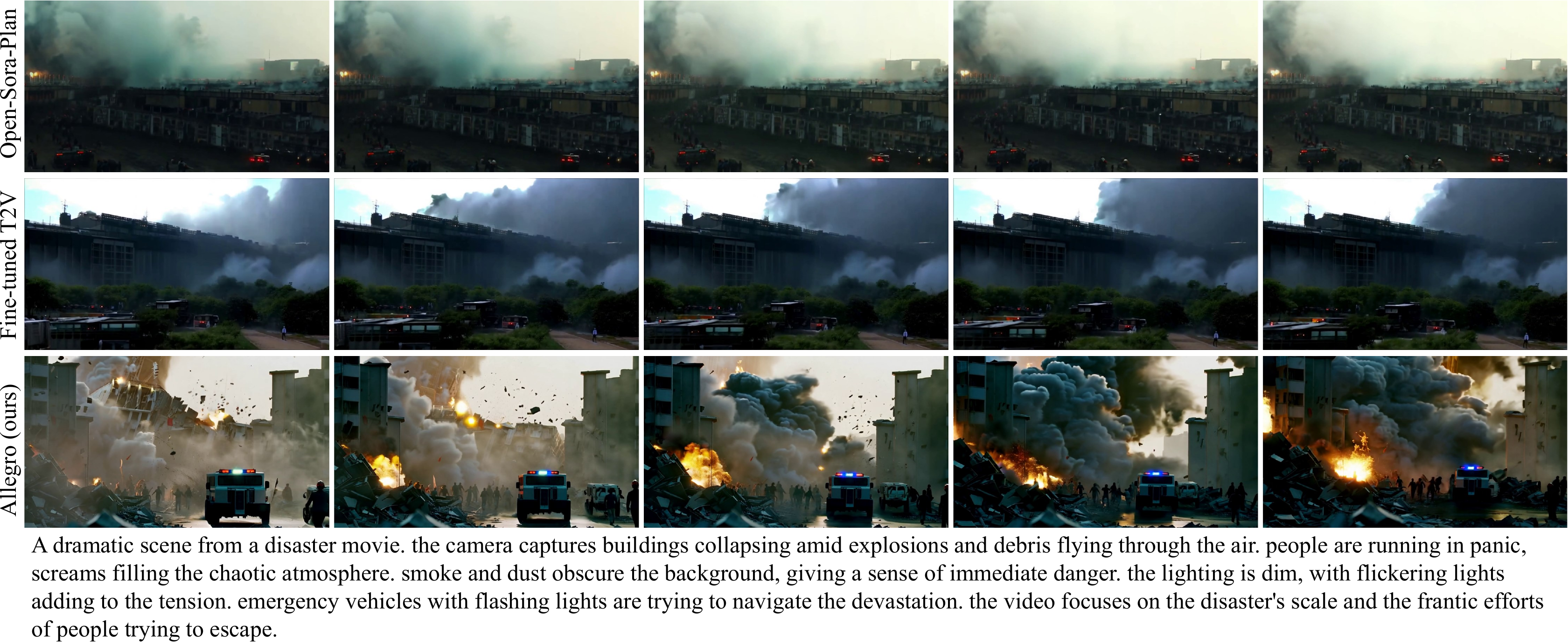}
    \caption{Qualitative comparison of generated frames from our Allegro and the Open-Sora-Plan model in text-to-video generation.}
    \label{fig:eval_textencoder_t2v}
\end{figure}

Some might speculate that the differences in performance are due to variations in data usage. To address this, we further fine-tuned the Open-Sora-Plan text-to-video model using the same dataset. As shown in the second row in Figure~\ref{fig:eval_textencoder_t2v}, even with extensive fine-tuning on high-quality data, the semantic modeling issues caused by using the mT5 text encoder in a fully English training dataset remain unresolved. From Figure~\ref{fig:eval_textencoder_t2i} and the third row in Figure~\ref{fig:eval_textencoder_t2v}, we can observe that replacing the text encoder with T5 and training as described in Section~\ref{sec:dit_training} on high-quality data leads to significant improvements in both text-to-image and text-to-video tasks.

The results also indicate that it is challenging to obtain a high-quality video model by directly fine-tuning the Open-Sora-Plan text-to-video model. This is primarily due to the inaccurate alignment between text and visual elements established during the initial text-to-image pre-training stage. Since the text-to-video training process mainly focuses on modeling motion, this issue is difficult to address through text-to-video fine-tuning alone.

\begin{figure}[t]
    \centering
    \includegraphics[width=\textwidth]{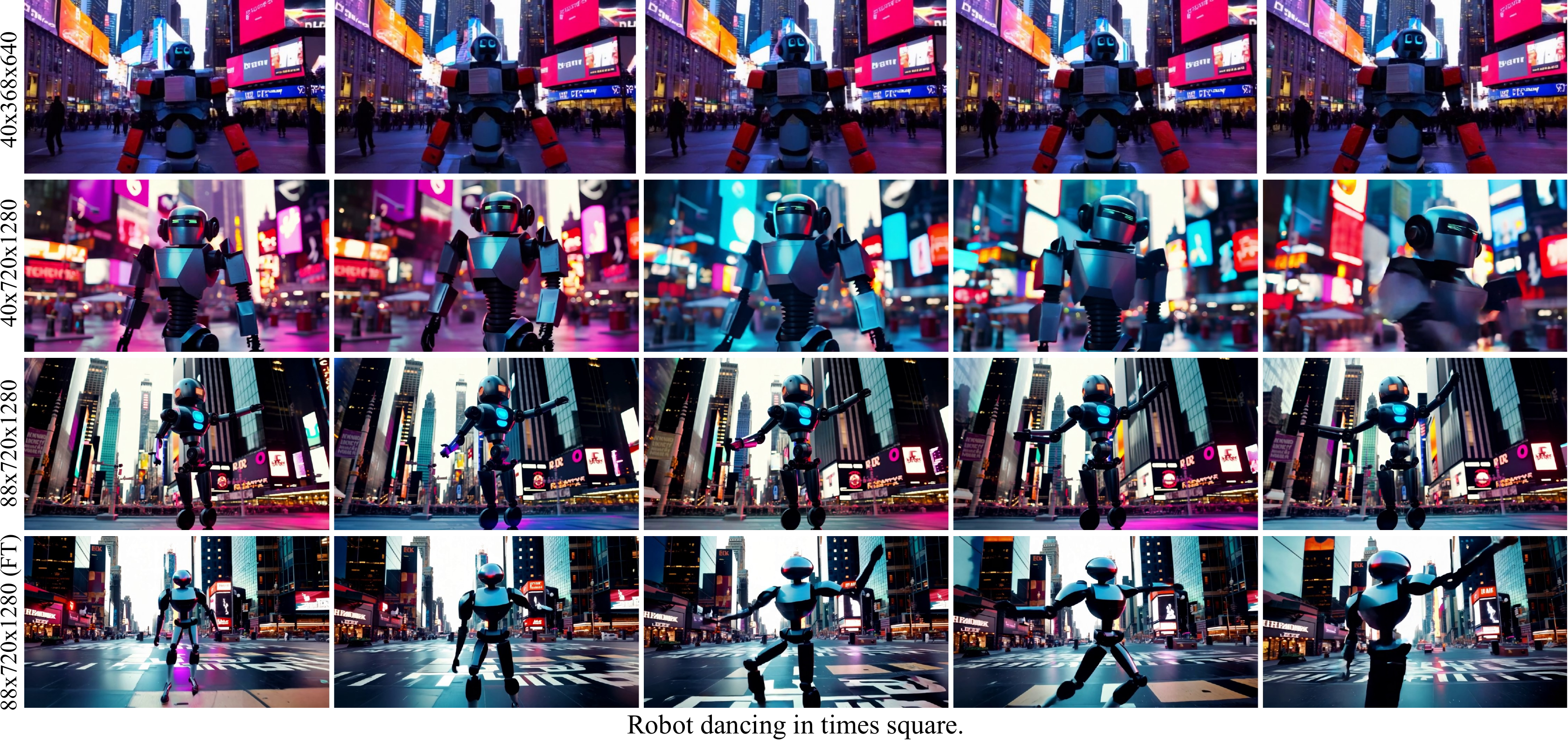}
    \caption{Qualitative comparison of generated video frames from our Allegro in different training stages.}
    \label{fig:eval_multistage}
\end{figure}

\textbf{Multi-Stage Training.}
As discussed in Section~\ref{sec:dit_training}, we train our model with a coarse-to-fine multi-stage training strategy. As shown in Figure~\ref{fig:eval_multistage}. In the initial $40\times 368\times 640$ stage, the model is initialized from the pre-trained text-to-image model and begins learning the continuity between video frames, resulting in relatively small-scale motion in the generated videos. Constrained by the VideoVAE's capacity for modeling at low resolutions, the overall video quality remains low, with noticeable jitter in the motion of finer details. In the subsequent $40\times 720 \times 1280$ stage, the increase in resolution significantly improves the visual quality of the generated videos. The model starts to better capture motion between frames, but due to the limited number of frames, there are still distortions and unnatural motion in the videos. Moving forward to the $88\times720\times1280$ stage, the increased frame count greatly enhances the model's ability to represent large-scale motion. Finally, by fine-tuning the model with a small amount of high-quality data at the same video size, the model's overall visual aesthetics, motion dynamics, and responsiveness to different types of text inputs are further improved.

\begin{figure}[t!]
    \centering
    \includegraphics[width=\textwidth]{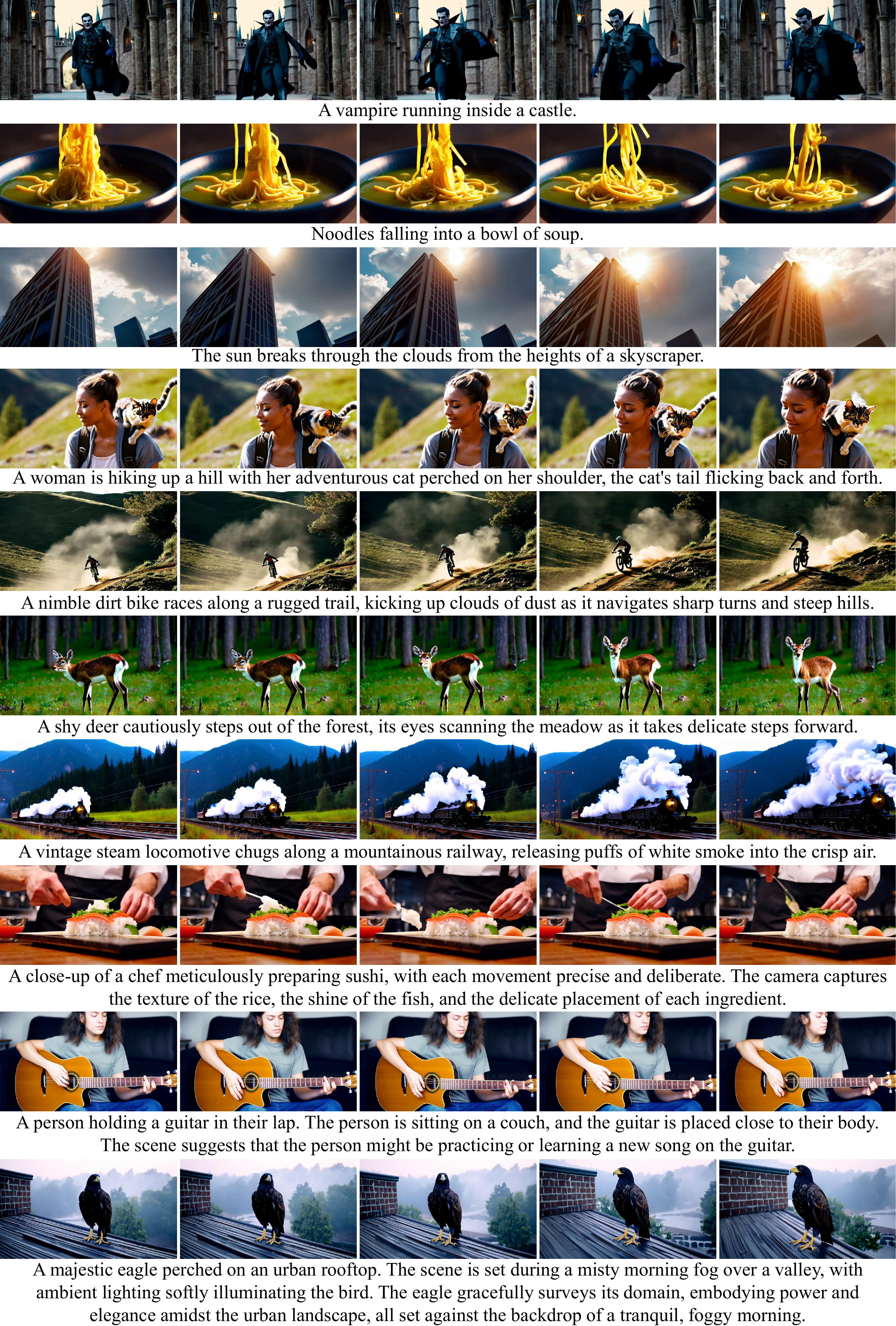}
    \caption{The generated frames of our Allegro with text inputs of varying lengths.}
    \label{fig:eval_vresult}
\end{figure}

\textbf{Visual Results.}
We show visual results of our proposed model, Allegro, in Figure~\ref{fig:eval_vresult}. 
These results demonstrate that our model is capable of handling text inputs of varying lengths while generating high-quality, dynamic videos. More visual results can be found in Appendix~\ref{sec:appx_moreresults}.
\section{More Thinkings}
\textbf{VideoVAE.}
Our VideoVAE, built on the architecture of a previous image VAE, incorporates numerous temporal CNN layers and applies overlapping and blending in both spatial and temporal dimensions, leading to extremely high computational costs. 

For future developments, it will be important to strike a balance between reconstruction quality and computational efficiency when designing the spatiotemporal modeling structure of VideoVAE. Additionally, to improve latent feature extraction, we should avoid spatial tiling and increase the spatiotemporal unit size during video encoding.

Under the same model parameter constraints and compression stride, increasing the latent channel size is a simple yet effective approach to enhancing the detail in reconstructed videos. However, reconstructing line-like semantics remains a challenging task for VideoVAE models that rely on RGB images. We believe that incorporating frequency domain features into the input of VideoVAE could potentially improve its ability to reconstruct line-like semantics in the future.

\textbf{VideoDiT.}
We have trained the current model with a sufficiently large dataset. However, as training progresses, improvements in certain areas have become limited. For example, the model demonstrates inadequate fine-grained instruction-following abilities, less natural interactions in more complex subject scenarios, and limited further enhancement in visual textures. Drawing from the experiences of LLMs and vision transformer models, increasing the model’s parameter count is a direct and promising approach to address these limitations.
For instance, expanding the model’s width enhances its capacity to capture more complex visual patterns. Increasing depth enables it to learn hierarchical representations. These two approaches are both crucial for understanding intricate subject interactions and generating more refined textures.

Although we have not performed comprehensive, large-scale experiments to explore the scaling limits in our model, we have conducted small-scale ablation studies. These experiments allow us to investigate the effects of increasing width and depth on a manageable scale, and the results suggest that further exploration of model scaling could yield promising improvements. Therefore, our findings offer some practical guidance for researchers and engineers in the open-source community who are interested in scaling up models. They can consider methods such as expanding the width and depth of their architectures, which have been proven effective in both the LLM and vision transformer domains. These insights can serve as a starting point for more in-depth, resource-intensive studies aimed at unlocking further advancements in model performance.

\textbf{Prompt Refiner.}
A general prompt refiner interprets, adjusts, or expands the given prompt to better match the model's capabilities; in other words, adjust the given prompt's distribution to match the distribution of training data captions. Considering that our training pipeline does not include training for text encoders, we argue that structuring the caption in training data can lower the complexity of training and lead to faster convergence. All the captions or prompts follow a specific manner, as mentioned in Section~\ref{sec:data_annotation}. Through structuralization, the general text-to-video task can be simplified to label-to-video. However, such a simplification inevitably harms generalization abilities, and that's why a prompt refiner is needed. 

For large-scale pretraining and better generalization ability, we suggest enhancing the training data's diversity and reducing the model's dependency on the refiner. To be specific, annotation methods, including leveraging multiple captioners with different linguistic styles, mixing the data of real use cases, and selecting the data based on human preference, are all good for generalization. This aspect of the work is left for future exploration.

\section{Future Work}
We are working on several features as our future works to enhance the capability of our proposed Allegro text-to-video model.

\textbf{Image-to-Video.} An image-to-video generation model with text condition, i.e. Text-Image-to-Video, can be built upon the foundation of a well-trained text-to-video model, adding the ability to incorporate visual context (images) alongside textual input. This enhancement allows the model to generate more accurate and visually rich videos by combining the descriptive power of text with the specific visual cues provided by images. The necessity of such a model arises from the limitations of text-only inputs, which can sometimes be ambiguous or lack detailed visual context. By integrating images, the model can create more precise, contextually relevant, and higher-quality videos, bridging the gap between user intent and the generated output. This is particularly valuable in commercial applications, where brands and businesses often need to convey specific visual themes or brand elements that may not be fully captured through text alone. Our text-image-to-video model is under training, based on our foundational text-to-video model, and will be released as another extended model of the Allegro family in the near future.

\textbf{Motion Control.} Under the combined introduction of text descriptions and appearance prior conditions, many video producers still have some more precise control requirements that are difficult to specify with text, among which the most notable is the motion control of the area of interest in the first frame.
Our motion control-related data has been preliminarily prepared, and the motion control model is currently being trained. We expect to release it subsequently to provide everyone with more fine-grained control options.

\section{Acknowledgement}

The authors appreciate Dongxu Li, Hanshu Yan, Haoning Wu, Huiguo He, Ruoyu Li, Wenhao Huang, Xin Yan, Yifan Ye, and Zixi Tuo (in alphabet order) for their valuable input and suggestions.

\bibliographystyle{abbrvnat}
\bibliography{main}

\appendix

\section{Data Analysis}
\label{sec:appx_data_analysis}
In this section, we conduct a statistical analysis of the training data, showing the statistical distribution across various dimensions. 
We compare the distribution differences in different training stages. Additionally, specific examples are shown to demonstrate how the data filtering process improves data quality.

\begin{figure}[h]
    \centering
    \begin{subfigure}[b]{0.3\textwidth}
        \centering
        \includegraphics[width=\textwidth]{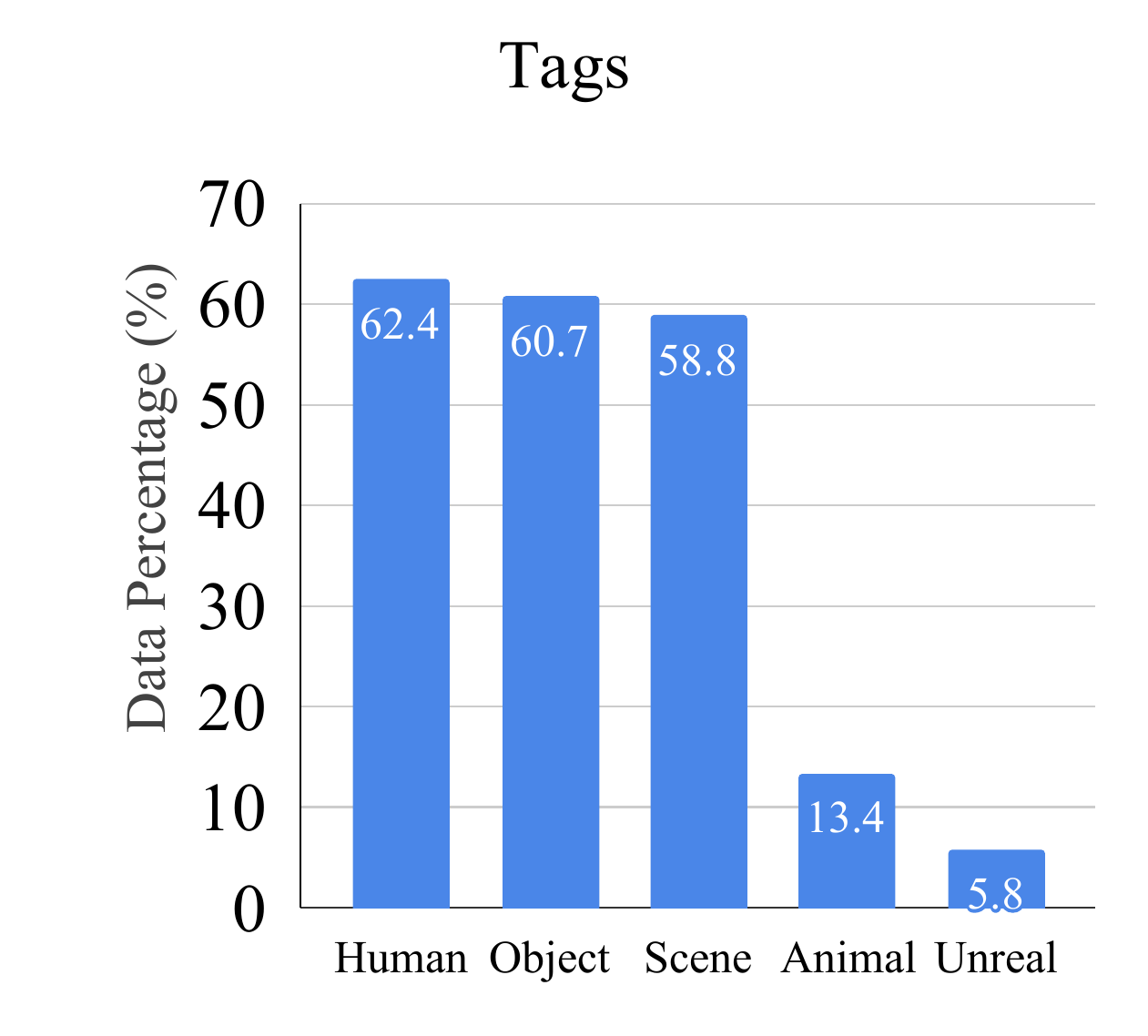}
        \caption{}
        \label{fig:appx_data_a}
    \end{subfigure}
    \begin{subfigure}[b]{0.3\textwidth}
        \centering
        \includegraphics[width=\textwidth]{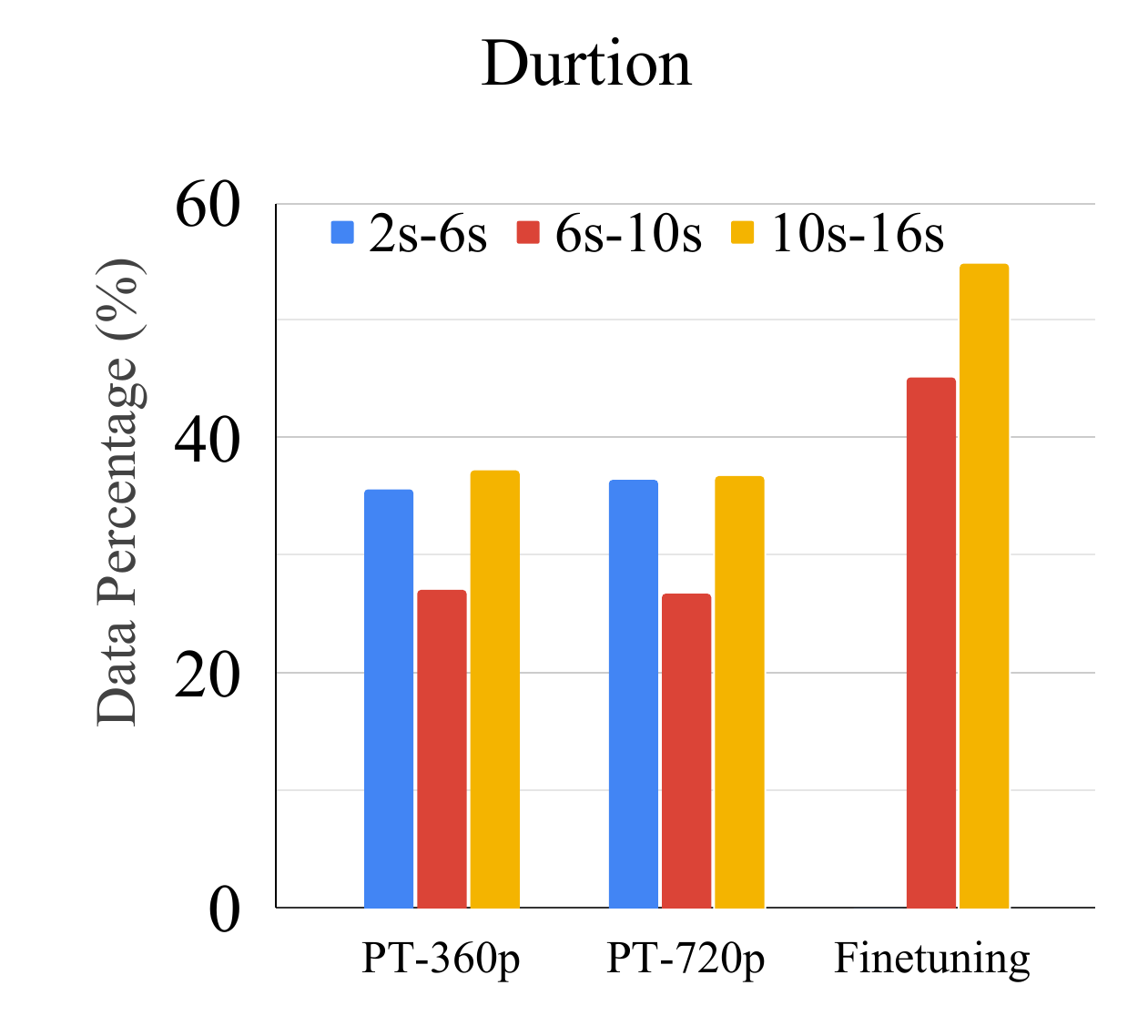}
        \caption{}
        \label{fig:appx_data_b}
    \end{subfigure}
    \begin{subfigure}[b]{0.3\textwidth}
        \centering
        \includegraphics[width=\textwidth]{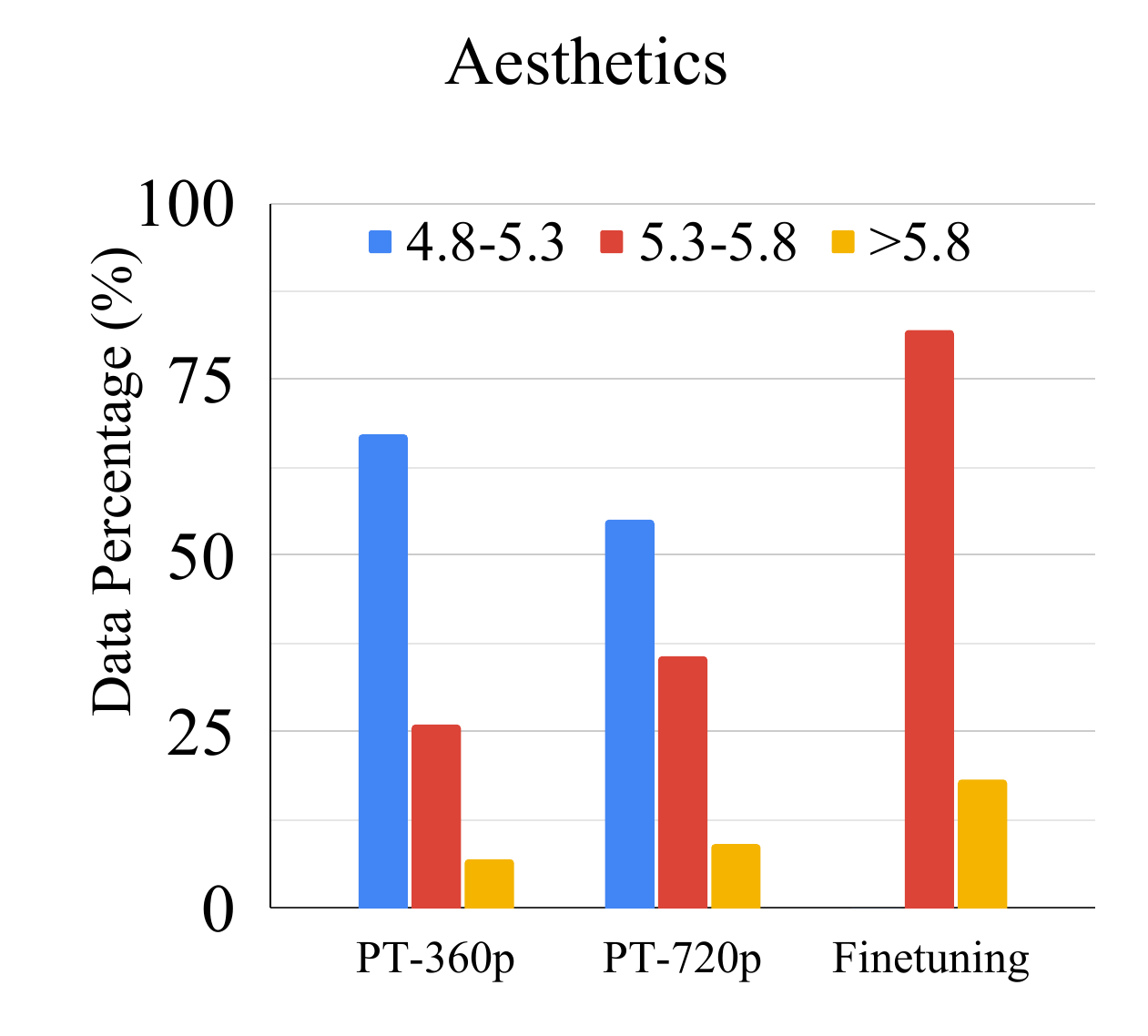}
        \caption{}
        \label{fig:appx_data_c}
    \end{subfigure}


    \begin{subfigure}[b]{0.3\textwidth}
        \centering
        \includegraphics[width=\textwidth]{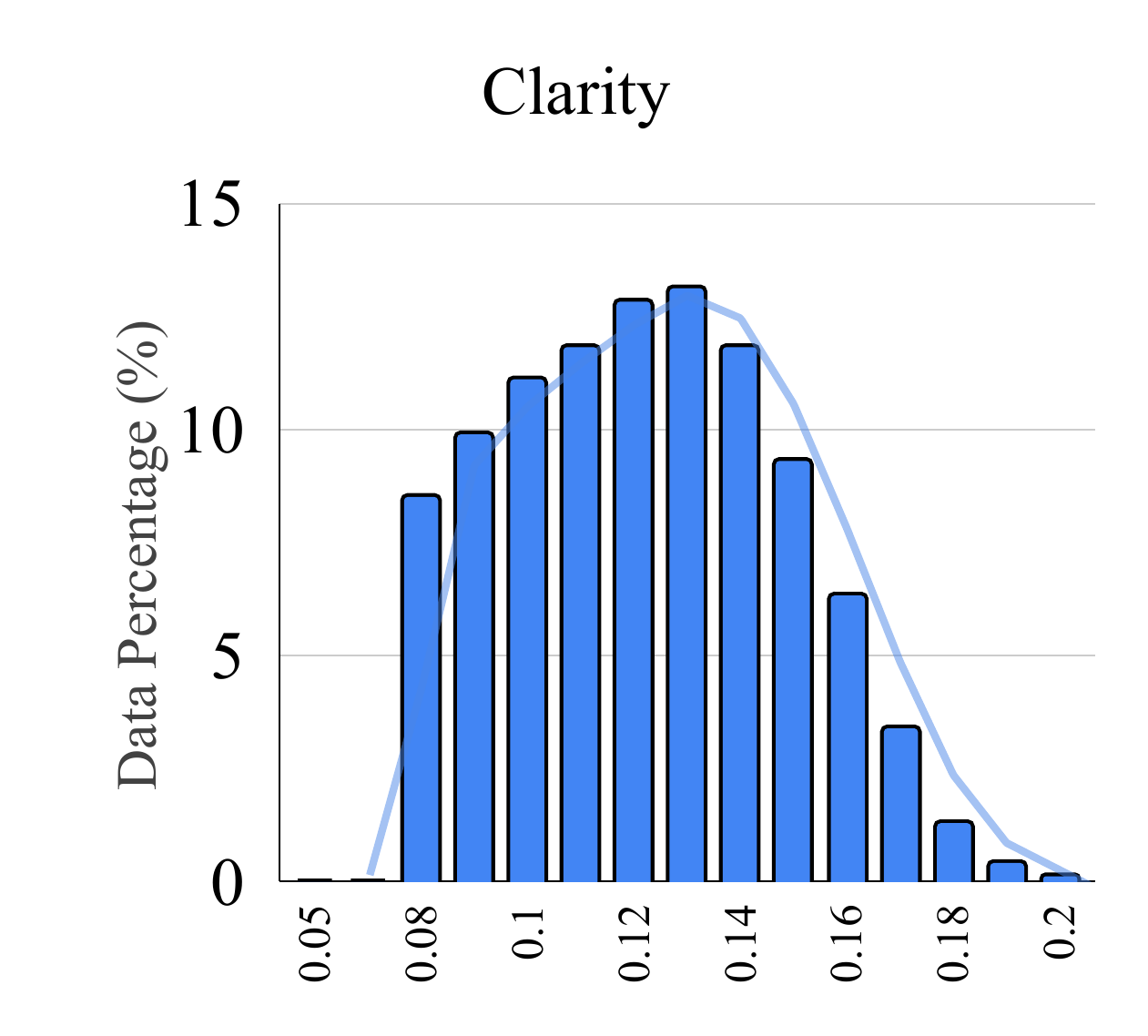}
        \caption{}
        \label{fig:appx_data_d}
    \end{subfigure}
    \begin{subfigure}[b]{0.3\textwidth}
        \centering
        \includegraphics[width=\textwidth]{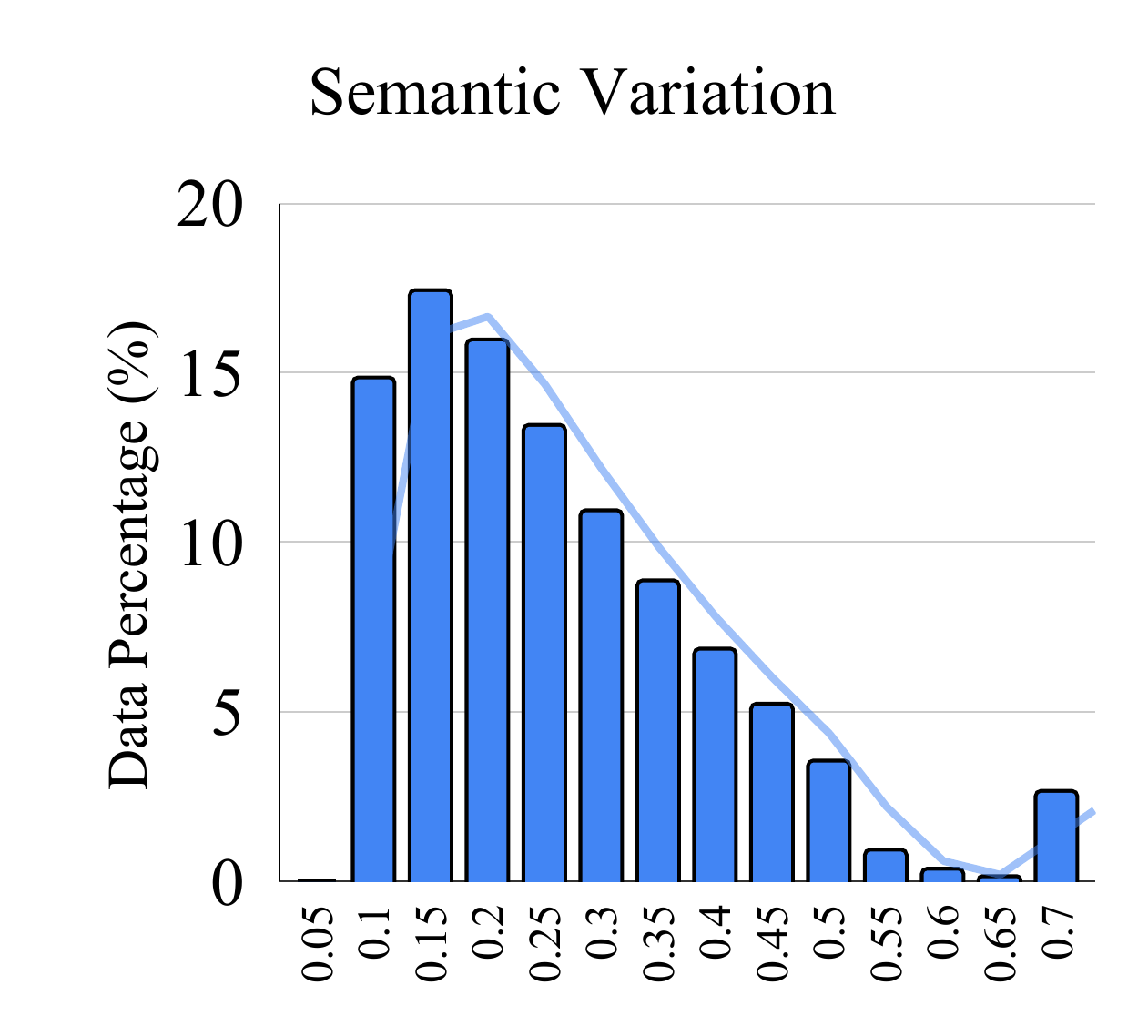}
        \caption{}
        \label{fig:appx_data_e}
    \end{subfigure}
    \begin{subfigure}[b]{0.3\textwidth}
        \centering
        \includegraphics[width=\textwidth]{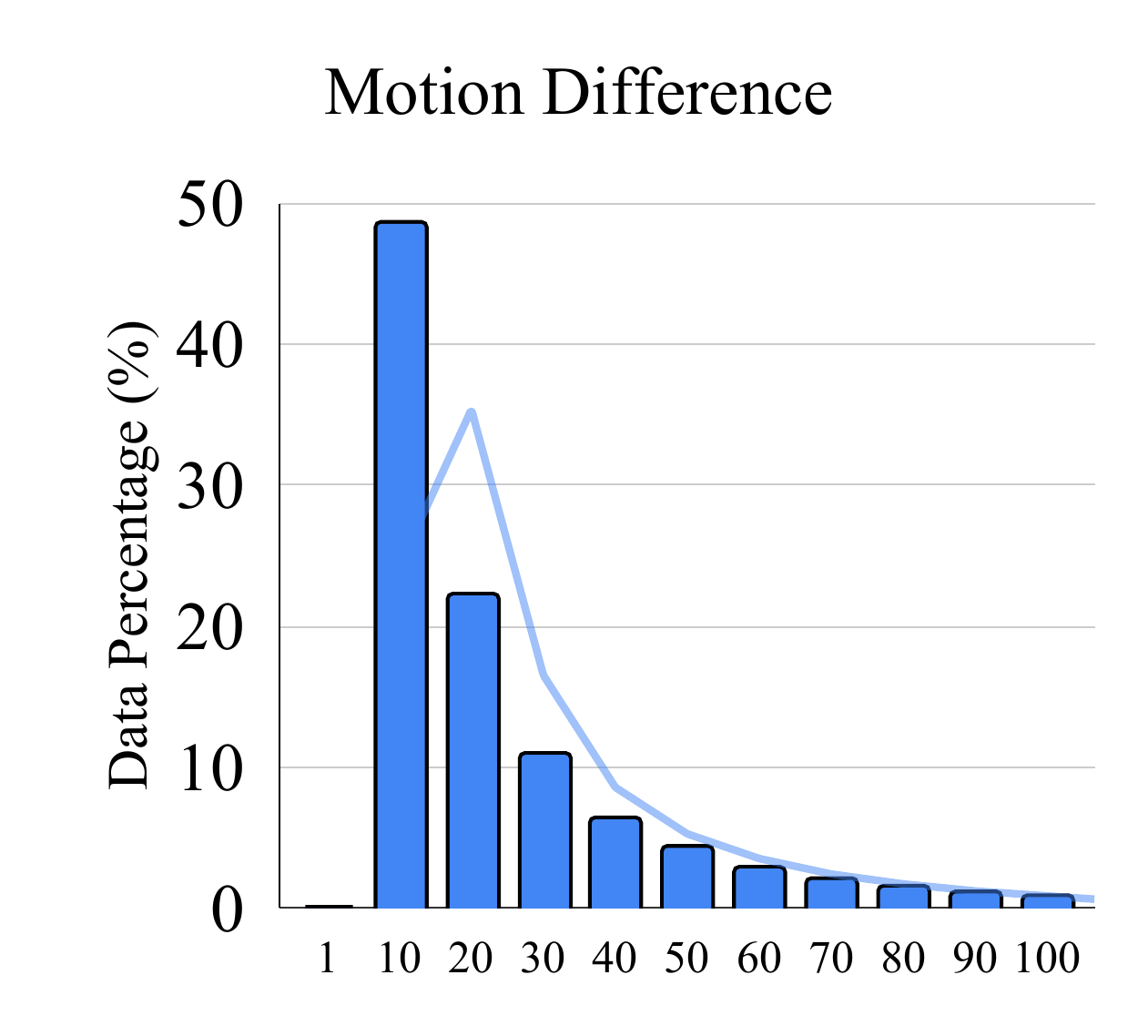}
        \caption{}
        \label{fig:appx_data_f}
    \end{subfigure}

    \caption{Data statistics in different dimensions. (a) Distribution of different tags in the training data. (b) and (c) shows the duration and aesthetic score of the data used in different training stages, respectively. (d)-(f) shows the data distributions of clarity, semantic variation, and motion difference, estimated with the scores of DOVER, LPIPS, and UniMatch, respectively.}
    \label{fig:appx_data}
\end{figure}

\textbf{Video duration, variation, and consistency.} 
Figure~\ref{fig:appx_data_b} shows how the data is distributed based on video duration across Text-to-Video training stages. 
The data are categorized into three duration buckets: 2s-6s (short-duration), 6s-10s (medium-duration), and 10s-16s (long-duration).
The pre-training stages show a balanced distribution across the three buckets in both 360p and 720p. In contrast, the fine-tuning stage shifts focus entirely to medium and long-duration videos, with no short-duration videos being used at this stage. 
In the early stages of training, the model focuses more on learning a large amount of semantic information. Single-frame images or videos with minimal motion and semantic changes play a more important role. 
As training progresses, there will be a greater demand for diversity in the richness of video content and the range of motion, which inevitably requires longer videos.
Additionally, it is also necessary to further quantify the degree of content variation and motion within the videos. 
Therefore, by introducing the evaluation metrics of the DOVER score, LPIPS score, and UniMatch score, it is possible to better control the selection of data that exhibits these characteristics. 
To improve data processing efficiency, these three metrics are only calculated during the fine-tuning stage. By jointly adjusting the thresholds of these three metrics, we obtained data with high clarity but moderate changes and motion during the fine-tuning stage.
The distribution of clarity, semantic variation, and motion are shown in Figure~\ref{fig:appx_data_d},~\ref{fig:appx_data_e} and~\ref{fig:appx_data_f}, respectively.
This suggests that longer videos with appropriate variations become more important as the model approaches final tuning to improve the model on long-duration video generation capability by this progressive training recipe.

\textbf{Resolution and Aesthetics.}
Figure~\ref{fig:appx_data_c} shows the distribution of aesthetic scores of the images/videos used during the Text-to-Video pre-training at 360p, 720p, and the fine-tuning stages.
Resolution during training impacts aesthetic quality but alone is not enough to achieve high levels of aesthetics. fine-tuning is crucial in boosting the model’s ability to generate more aesthetically pleasing content. To achieve higher-quality aesthetics in video generation, incorporating a fine-tuning step after pre-training at higher resolutions (like 720p) is essential. This ensures the model learns to focus not just on structural aspects but on visual appeal. Aesthetic filtering can help in curating training datasets for fine-tuning, ensuring the model receives feedback on what constitutes high-quality, aesthetically pleasing video content.

\begin{figure}[t]
    \centering
    \includegraphics[width=0.9\textwidth]{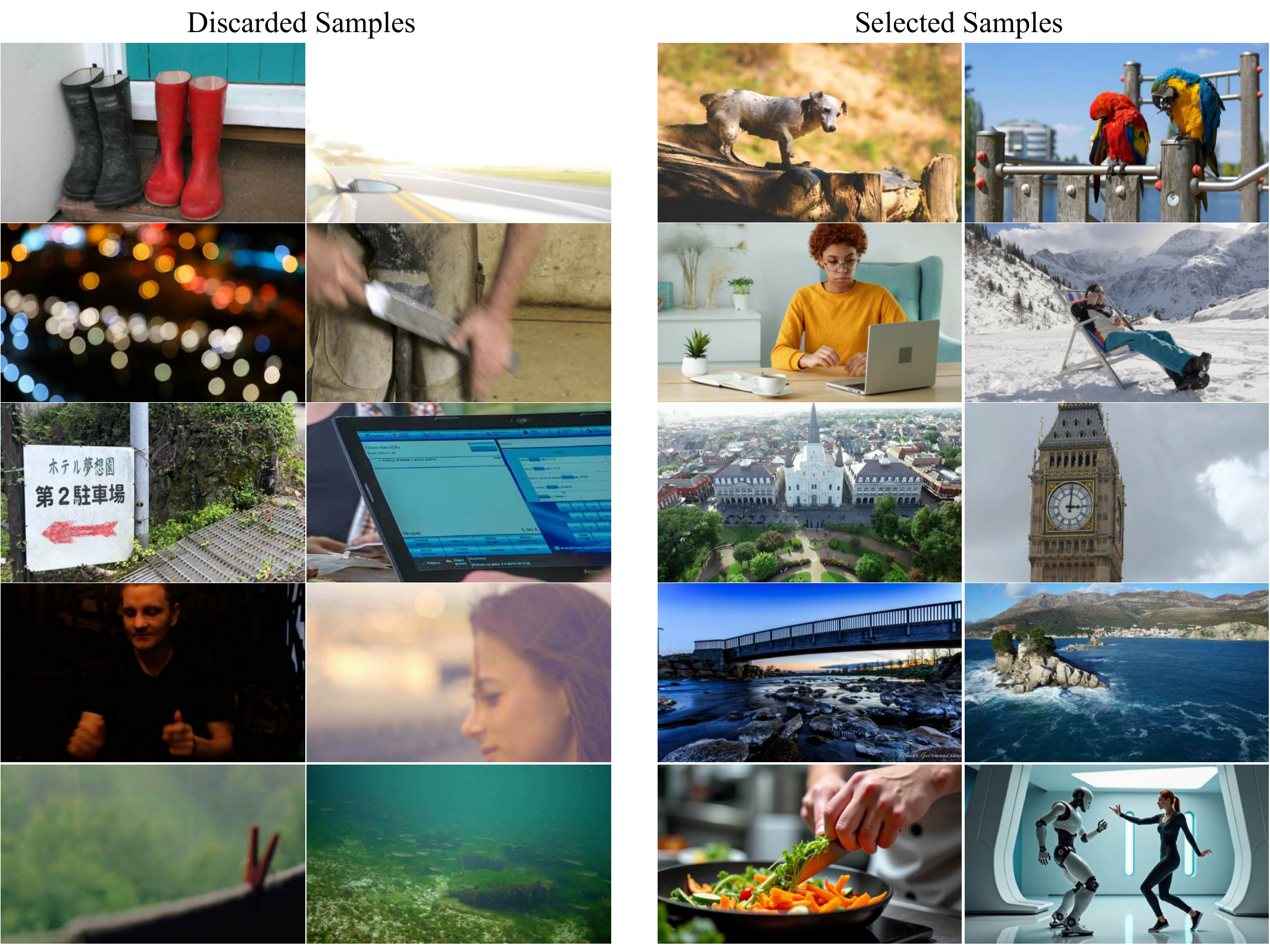}
    \caption{The selected and discarded data samples by the data filtering. The discarded samples include blurry samples, poorly focused, large text areas, or lack of clear subject matter, such as a pair of boots, unfocused city lights, and random objects. The selected samples are sharper, more aesthetically appealing, and have well-defined subjects, including vibrant scenes of nature, architecture, and human activities like cooking or working on a laptop.}
    \vspace{-2ex}
    \label{fig:appx_train_sample}
\end{figure}

We provide some training examples from different training stages to visually demonstrate the differences in data quality before and after the data filtering pipeline, shown in Figure~\ref{fig:appx_train_sample}. 
The discarded samples include blurry samples, poorly focused, large text areas, or lack of clear subject matter, such as a pair of boots, unfocused city lights, and random objects. The selected samples are sharper, more aesthetically appealing, and have well-defined subjects, including vibrant scenes of nature, architecture, and human activities like cooking or working on a laptop. The overall quality and clarity of the selected samples are visibly superior.

\textbf{Tags and Annotations}
Figure~\ref{fig:appx_data_a} shows the proportion of different tags in the training data, which are derived from the coarse-grained captions obtained during data filtering.
We can observe that, due to the natural distribution of the data, content related to people, objects, and landscapes occupies the majority. On the one hand, this helps the model better understand the semantics and motion patterns of the physical world by aligning with real-world data distributions. On the other hand, it enables the model to more effectively respond to instructions that are close to real-life scenarios, which is a crucial capability for a commercial-level video generation model.

\begin{figure}[t!]
    \centering
    \includegraphics[width=\textwidth]{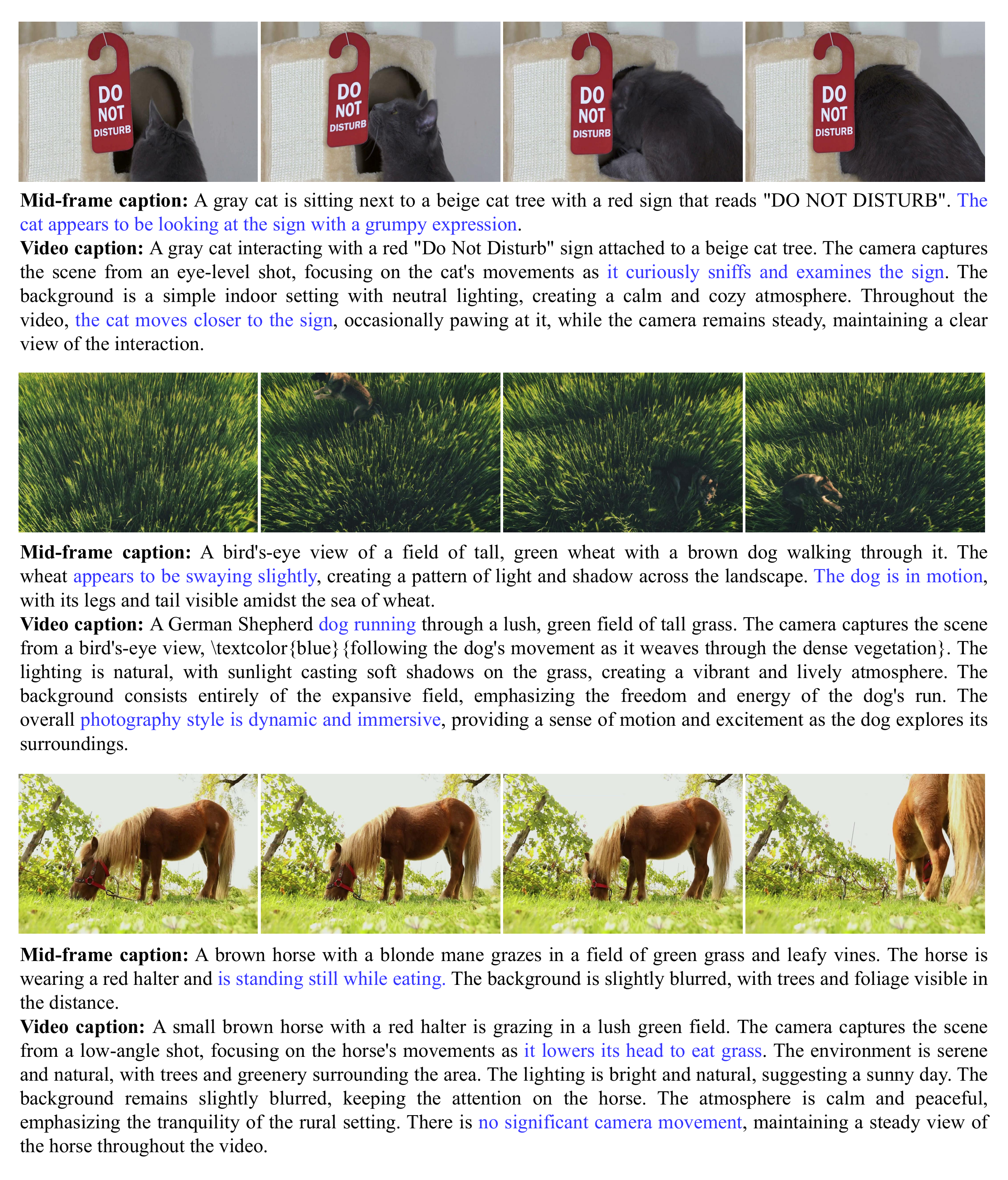}
    \caption{Fine-grained captions for the middle frame and video, sampled from training data. The blue text in the captions highlights the difference between the two types of captions. The \textbf{mid-frame caption} focuses more on fine-grained spatial information, while the missing temporal information is provided by the \textbf{video caption}.}
    \vspace{-2ex}
    \label{fig:appx_data_annotation}
\end{figure}

As mentioned in Section \hyperref[sec:data_annotation]{2.1.2}, for each video sample, we provide two captions: one for the middle frame and one for the entire video. The former tends to describe spatial information, while the latter incorporates temporal changes. This helps the model achieve fine-grained language-to-visual mapping across both spatial and temporal dimensions. Meanwhile, we use a video captioner based on Aria, which provides multi-level information, including foreground, background, style, environment, and camera movement. This offers the model potential capabilities to handle multi-granularity control, which is often required in commercial environments at the data annotation level. Figure~\ref{fig:appx_data_annotation} shows some examples of fine-grained captions for the middle frame and video in the training data.

\section{Model Specifications}
\label{sec:appx_modelspec}
The model parameter settings of Allegro are illustrated in Table~\ref{tab:model_param}.

\begin{table}[h]
\begin{tabular}{lc||lc}
\hline
Param. Name     & Param. Value  & Param. Name    & Param. Value       \\ \hline
in\_channels            & 4             & caption\_channels      & 4096              \\
out\_channels           & 4             & cross\_attention\_dim  & 2304              \\
num\_layers             & 32            & attention\_head\_dim   & 96                \\
interpolation\_scale\_h & 2.0           & num\_attention\_heads  & 24                \\
interpolation\_scale\_t & 2.2           & attention\_bias        & true              \\
interpolation\_scale\_w & 2.0           & dropout                & 0.0               \\
patch\_size             & 2             & activation\_fn         & gelu-approximate  \\
patch\_size\_t          & 1             & norm\_type             & ada\_norm\_single \\
sample\_size            & {[}90, 160{]} & norm\_num\_groups      & 32                \\
sample\_size\_t         & 22            & norm\_eps              & 1e-06             \\
use\_rope               & true          & num\_embeds\_ada\_norm & 1000              \\ \hline
\end{tabular}
\vspace{1ex}
\caption{The parameter settings of Allegro.}
\label{tab:model_param}
\vspace{-2ex}
\end{table}

\section{More Results on VBench}
\label{sec:appx_vbench}
\begin{figure}[h]
    \centering
    \includegraphics[width=0.7\textwidth]{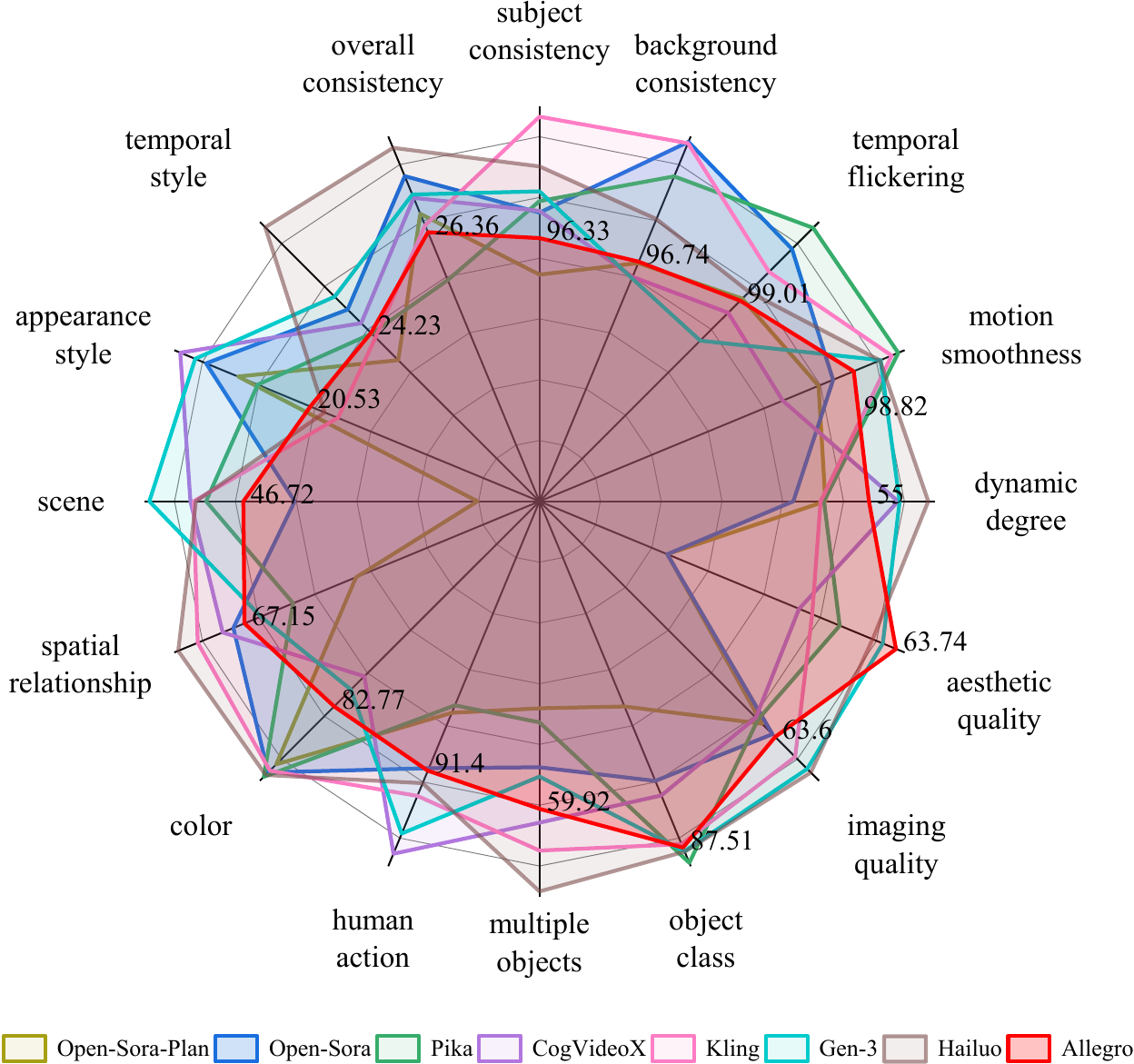}
    \caption{Comparison of Allegro and other video generation models on VBench.}
    \label{fig:appx_vbench}
\end{figure}

We show all dimensions of VBench in Figure~\ref{fig:appx_vbench}.

\section{More Visual Comparisions}
\label{sec:appx_morecmp}
\begin{figure}[t]
    \centering
    \includegraphics[width=\textwidth]{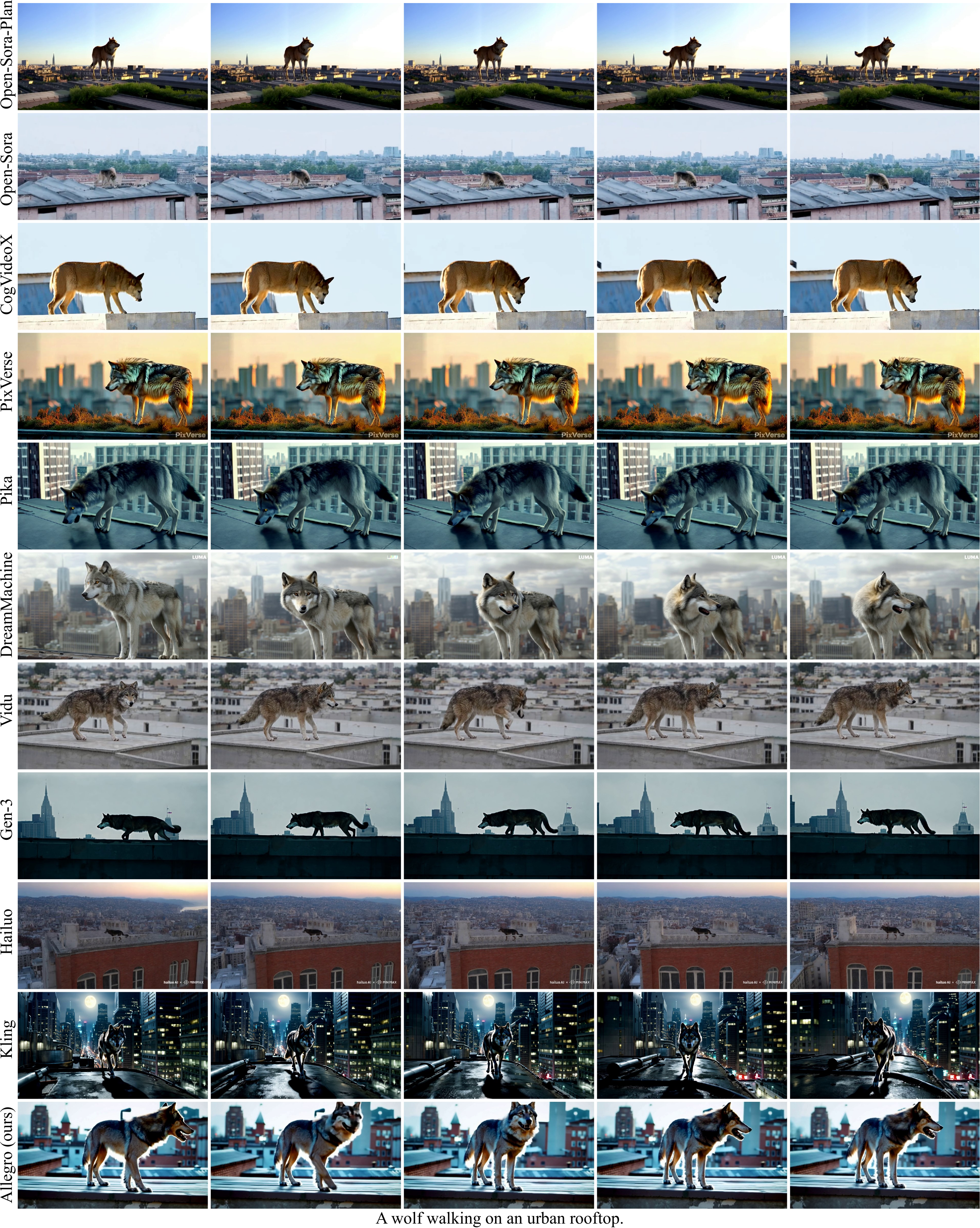}
    \caption{Qualitative comparisons of our Allegro with SOTA methods for the user study.}
    \label{fig:appx_userstudy2}
\end{figure}

\begin{figure}[t]
    \centering
    \includegraphics[width=\textwidth]{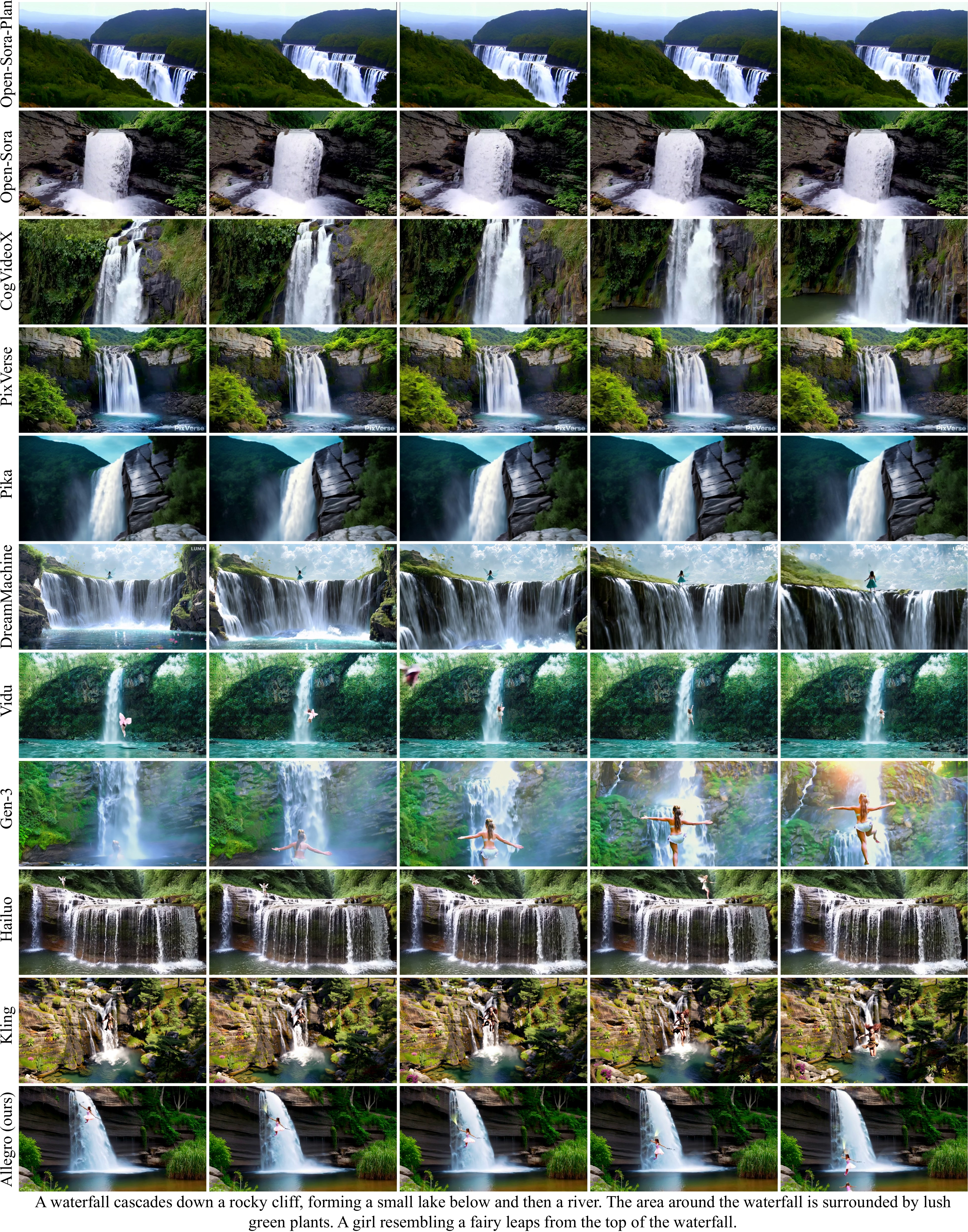}
    \caption{Qualitative comparisons of our Allegro with SOTA methods for the user study.}
    \label{fig:appx_userstudy3}
\end{figure}

\begin{figure}[t]
    \centering
    \includegraphics[width=\textwidth]{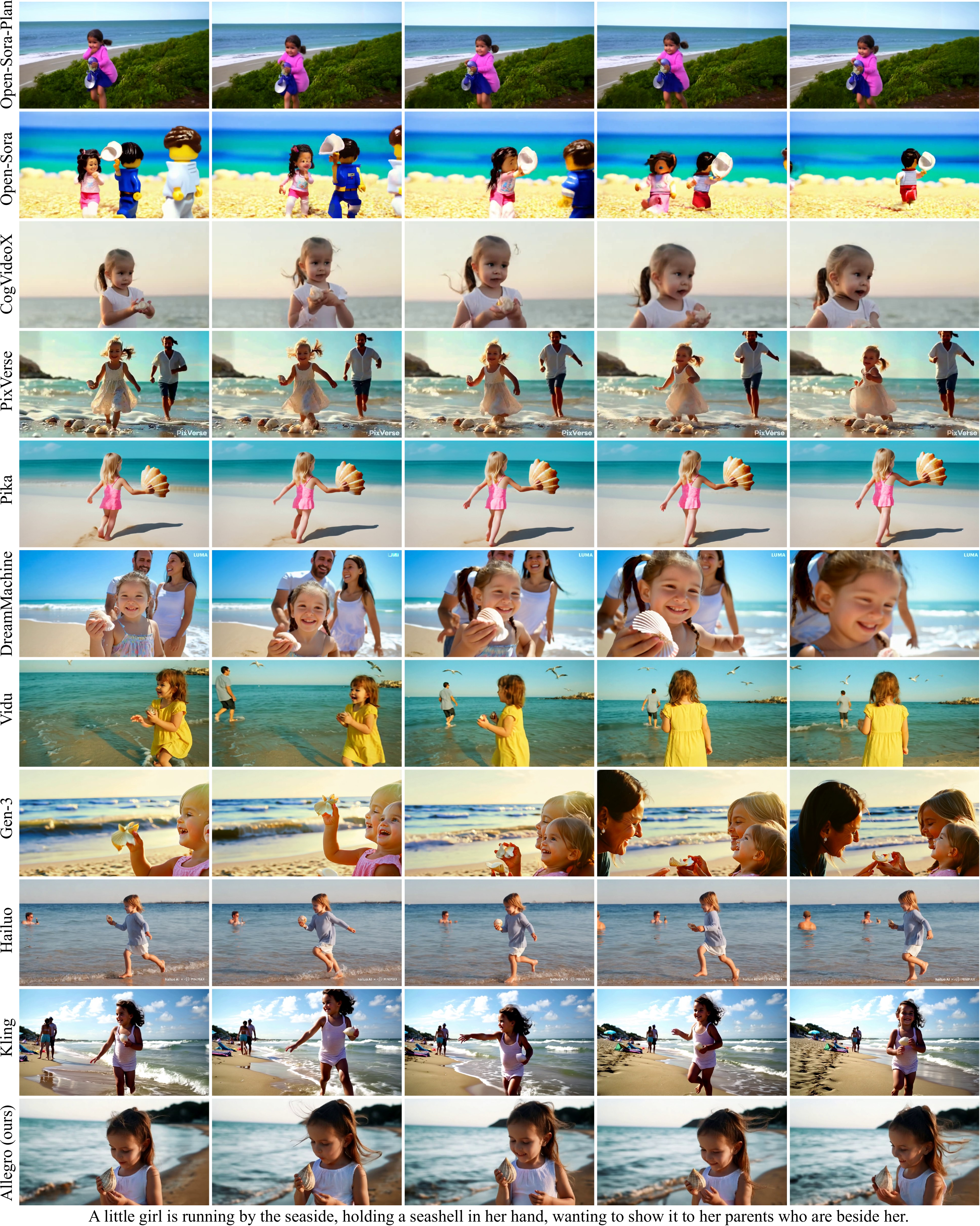}
    \caption{Qualitative comparisons of our Allegro with SOTA methods for the user study.}
    \label{fig:appx_userstudy4}
\end{figure}

\begin{figure}[t]
    \centering
    \includegraphics[width=\textwidth]{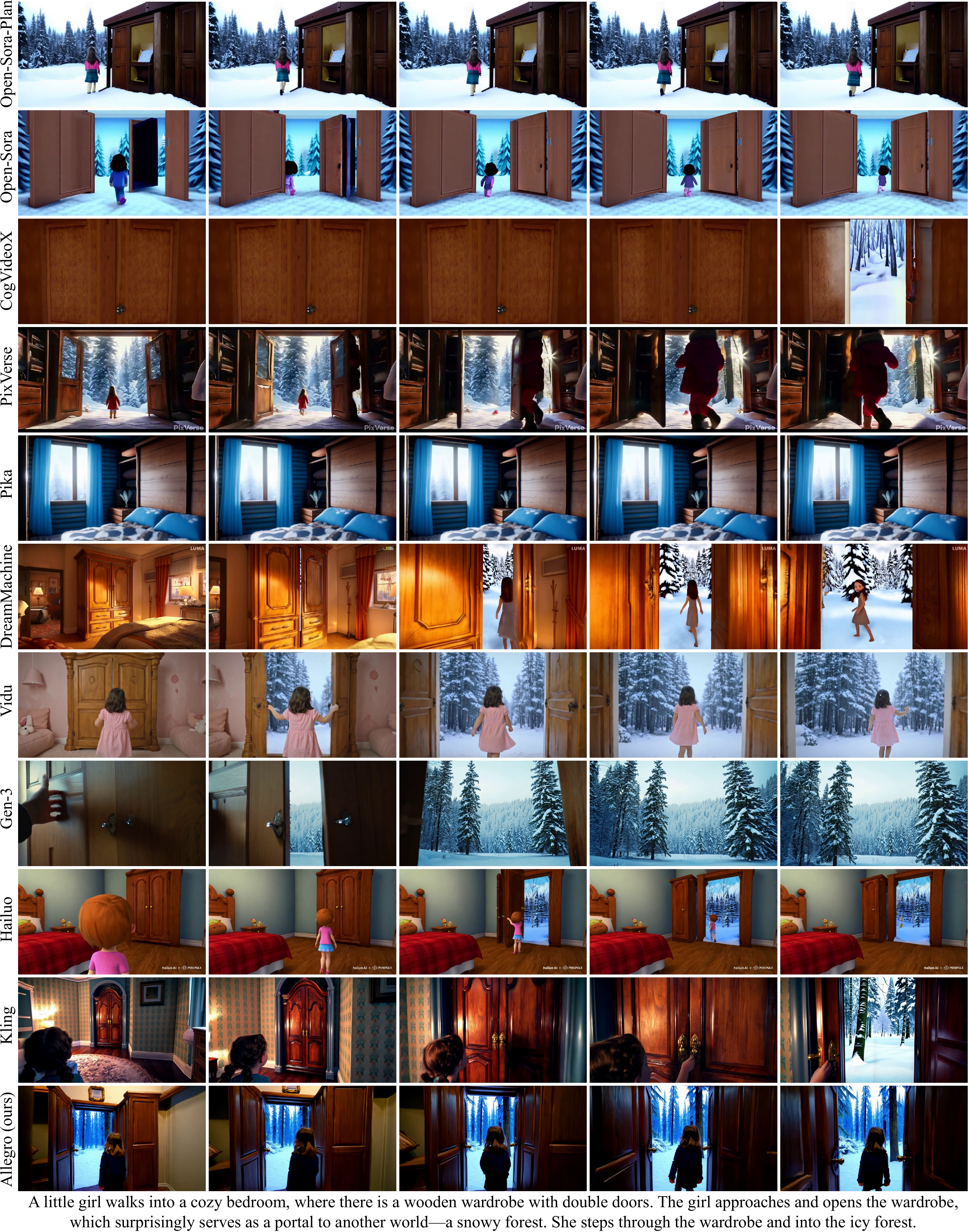}
    \caption{Qualitative comparisons of our Allegro with SOTA methods for the user study.}
    \label{fig:appx_userstudy5}
\end{figure}

We show more visual results of our method and other methods in Figure~\ref{fig:appx_userstudy2}, Figure~\ref{fig:appx_userstudy3}, Figure~\ref{fig:appx_userstudy4} and Figure~\ref{fig:appx_userstudy5}.

\section{More Visual Results}
\label{sec:appx_moreresults}
\begin{figure}[t]
    \centering
    \includegraphics[width=\textwidth]{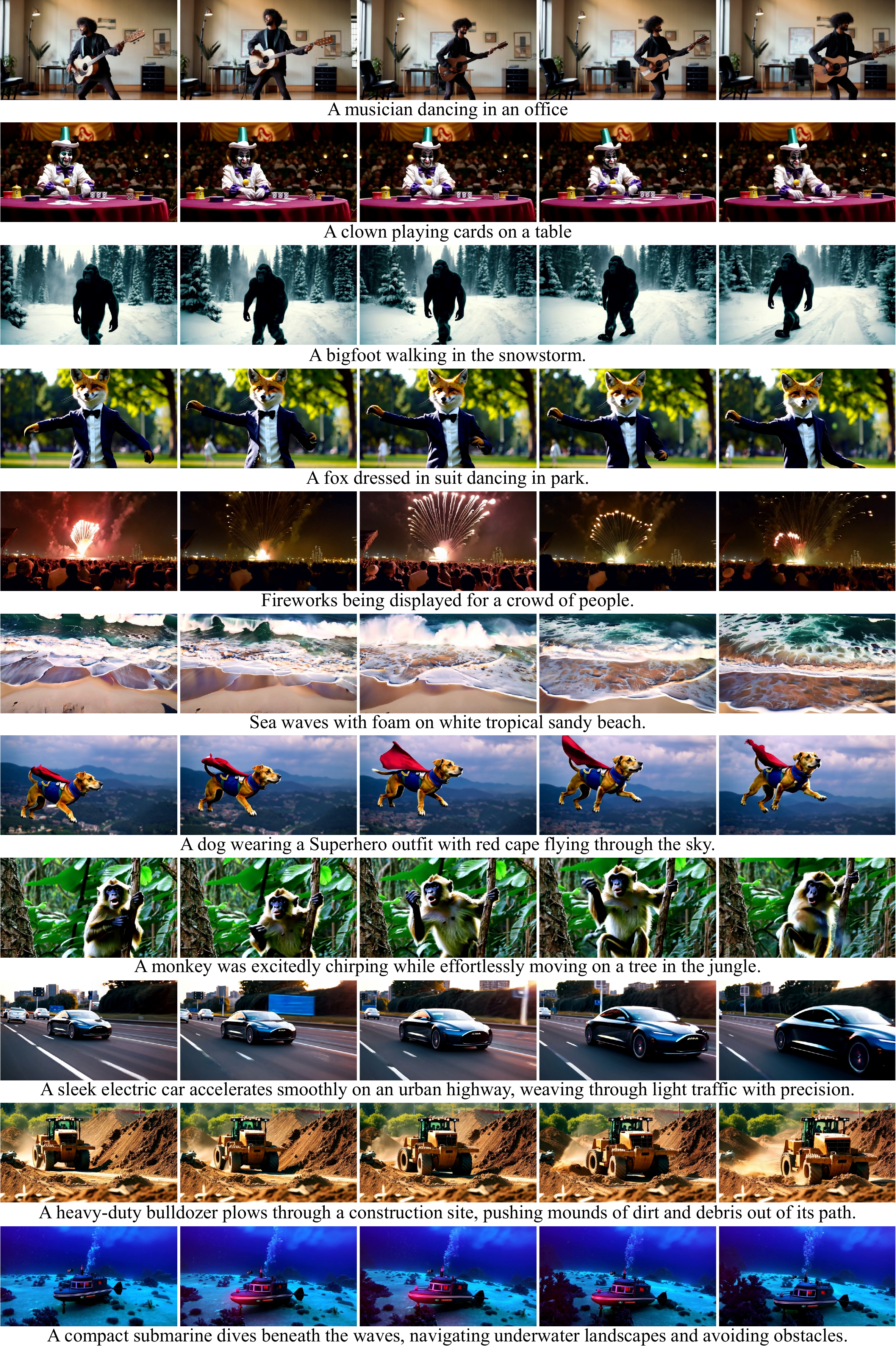}
    \caption{The generated video frames of our Allegro with text inputs of varying lengths.}
    \label{fig:eval_vresult2}
\end{figure}

\begin{figure}[t]
    \centering
    \includegraphics[width=\textwidth]{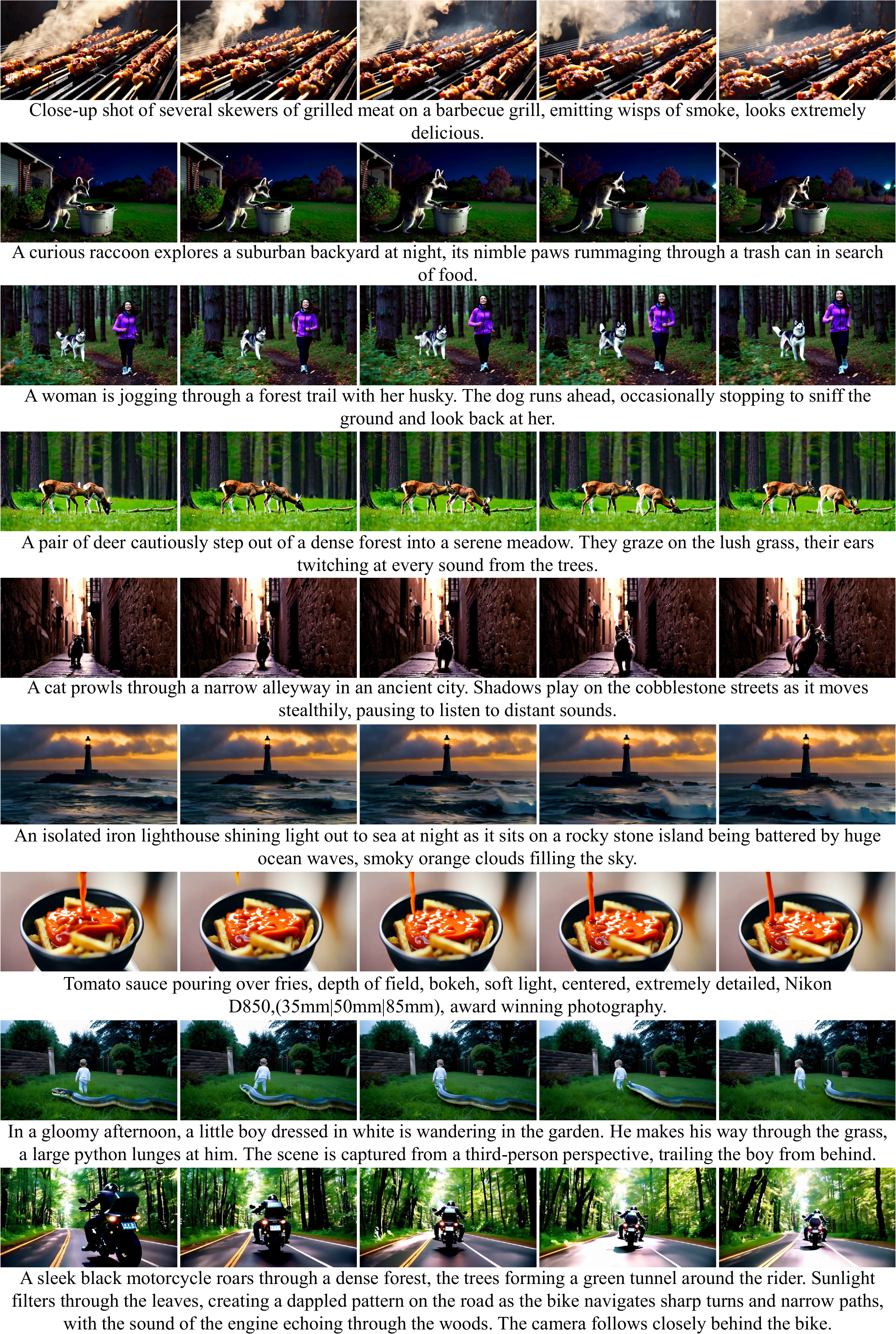}
    \caption{The generated video frames of our Allegro with text inputs of varying lengths.}
    \label{fig:eval_vresult3}
\end{figure}

\begin{figure}[t]
    \centering
    \includegraphics[width=\textwidth]{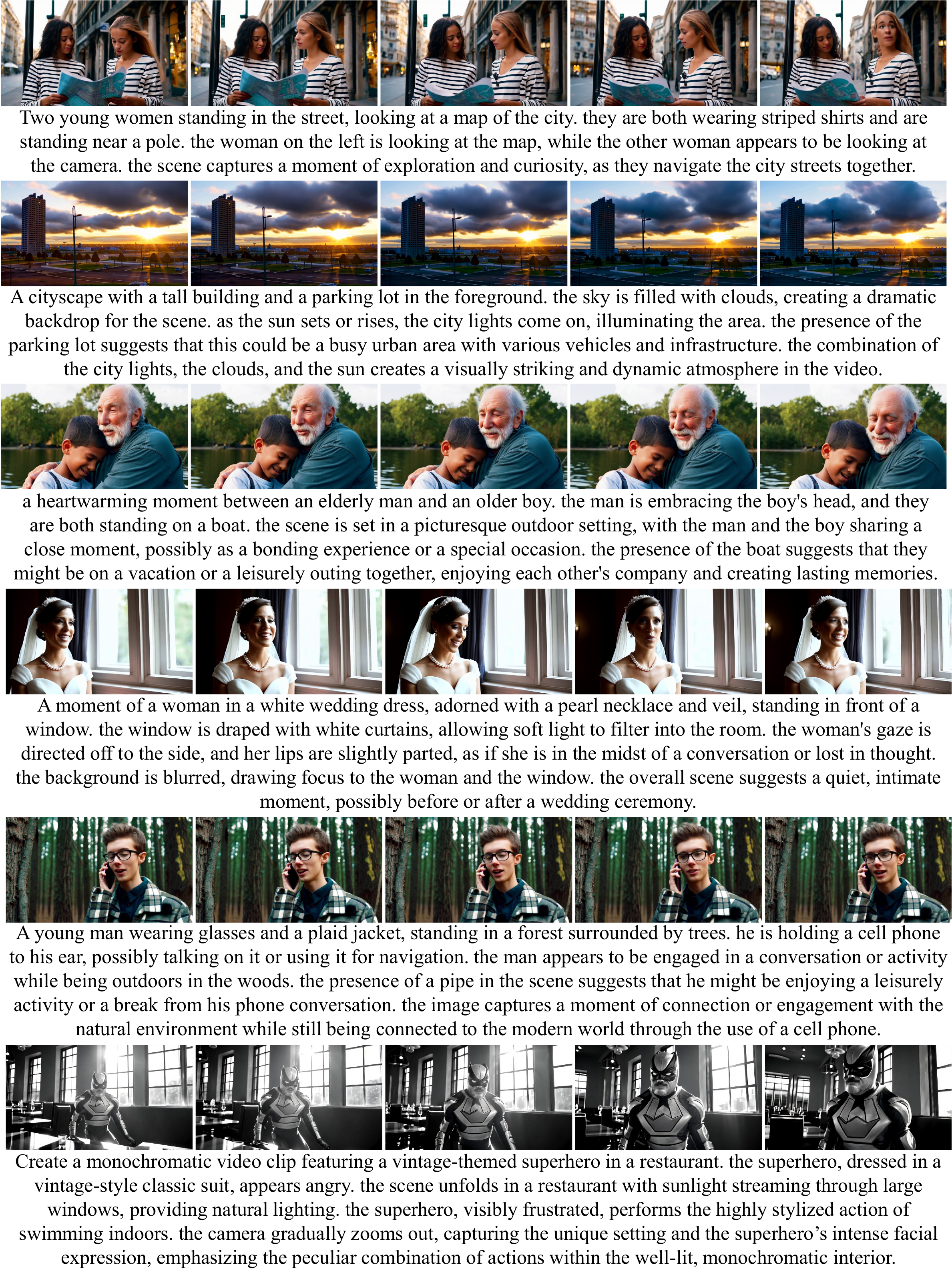}
    \caption{The generated video frames of our Allegro with text inputs of varying lengths.}
    \label{fig:eval_vresult4}
\end{figure}

We show more visual results of our Allegro in Figure~\ref{fig:eval_vresult2}, Figure~\ref{fig:eval_vresult3} and Figure~\ref{fig:eval_vresult4}.

\end{document}